\documentclass[review, letterpaper, compsoc]{IEEEtran}

\usepackage{amsmath,amsfonts,amssymb,mathtools,bm}
\usepackage{graphicx,epsfig,epstopdf,rotating}
\usepackage{algorithm,algpseudocode}
\usepackage{color,verbatim}
\usepackage{multirow,booktabs}
\usepackage[pagebackref=false,breaklinks=true,letterpaper=true,colorlinks,linkcolor=blue,urlcolor=blue,citecolor=blue,bookmarks=false]{hyperref}
\usepackage{lineno}

\newcommand{\etal}{\textit{et al}.}
\renewcommand{\v}[1]{\ensuremath{\mathbf{#1}}}

\DeclareMathOperator*{\argmin}{arg\,min}
\DeclareMathOperator*{\argmax}{arg\,max}
\begin{document}

\title{Online Localization and Prediction of \\Actions and Interactions}

\author{Khurram~Soomro,~\IEEEmembership{Member,~IEEE,}
        Haroon~Idrees,~\IEEEmembership{Member,~IEEE,}
        and~Mubarak~Shah,~\IEEEmembership{Fellow,~IEEE}% <-this % stops a space
\IEEEcompsocitemizethanks{\IEEEcompsocthanksitem K. Soomro, H. Idrees and M. Shah are with the Center for Research in Computer Vision (CRCV), University of Central Florida, Orlando, FL, 32816. E-mail: \{ksoomro,haroon,shah\}@eecs.ucf.edu}}

\markboth{IEEE TRANSACTIONS ON PATTERN ANALYSIS AND MACHINE INTELLIGENCE}%
{Shell \MakeLowercase{\textit{et al.}}: Bare Demo of IEEEtran.cls for Computer Society Journals}

\IEEEcompsoctitleabstractindextext{
\begin{abstract}
This paper proposes a person-centric and online approach to the challenging problem of localization and prediction of actions and interactions in videos. Typically, localization or recognition is performed in an offline manner where all the frames in the video are processed together. This prevents timely localization and prediction of actions and interactions - an important consideration for many tasks including surveillance and human-machine interaction.
\\
%After localizing
%actions and interactions at each time-step (frame), we refine
%poses in a batch of few frames by imposing consistency
%in locations and appearance of joints as well as scale of
%poses. Once the pose has been estimated and refined at
%current time-step, the superpixel-based appearance model
%is updated to avoid visual drift. This process is repeated
%for every frame in an online manner and gives
%human localization at every frame.

In our approach, we estimate human poses at each frame and train discriminative appearance models using the superpixels inside the pose bounding boxes. Since the pose estimation per frame is inherently noisy, the conditional probability of pose hypotheses at current time-step (frame) is computed using pose estimations in the current frame and their consistency with poses in the previous frames.
%we impose smoothness constraints across time on the joints to improve pose estimation.
Next, both the superpixel and pose-based foreground likelihoods are used to infer the location of actors at each time through a Conditional Random Field enforcing spatio-temporal smoothness in color, optical flow, motion
boundaries and edges among superpixels. The issue of visual drift is handled by updating the appearance models, and refining poses using motion smoothness on joint locations, in an online manner. For online prediction of action (interaction) confidences, we propose an approach
%uses that dynamic programming on SVM scores obtained on short segments of the video, thereby capturing sequential information of the actions. An alternate faster approach is
based on Structural SVM that operates on short video segments, and is trained with the objective that confidence of an action or interaction increases as time progresses. Lastly, we quantify the performance of both detection and prediction together, and analyze how the prediction accuracy varies as a time function of observed action (interaction) at different levels of detection performance. Our experiments on several datasets suggest that despite using only a few frames to localize actions (interactions) at each time instant, we are able to obtain competitive results to state-of-the-art \textit{offline} methods.
\end{abstract}

}

\maketitle

\IEEEdisplaynotcompsoctitleabstractindextext \IEEEpeerreviewmaketitle

% pami abstract: This paper proposes a person-centric and online approach to the challenging problem of localization and prediction of actions and interactions in videos which is typically performed in an offline manner. In our approach, we estimate human poses at each frame and train discriminative appearance models using the superpixels inside the pose bounding boxes. Since the pose estimation is inherently noisy, the conditional probability of pose hypotheses at current time-step is computed using pose estimations in the current frame and their consistency with poses in the previous frames. Next, both the superpixel and pose-based foreground likelihoods are used to infer the location of actors at each time through a CRF. The issue of visual drift is handled by updating the appearance models and refining poses in an online manner. For online prediction of action (interaction), we propose an approach based on Structural SVM that operates on short video segments. Lastly, we quantify the performance of both detection and prediction together. Our experiments on several datasets suggest that despite using only a few frames to localize actions (interactions) at each time instant, we are able to obtain competitive results to state-of-the-art `offline' methods.

%%%%%%%%% INTRODUCTION %%%%%%%%%
\section{Introduction}
\begin{figure}[t]
\centering
\includegraphics[width=1\columnwidth]{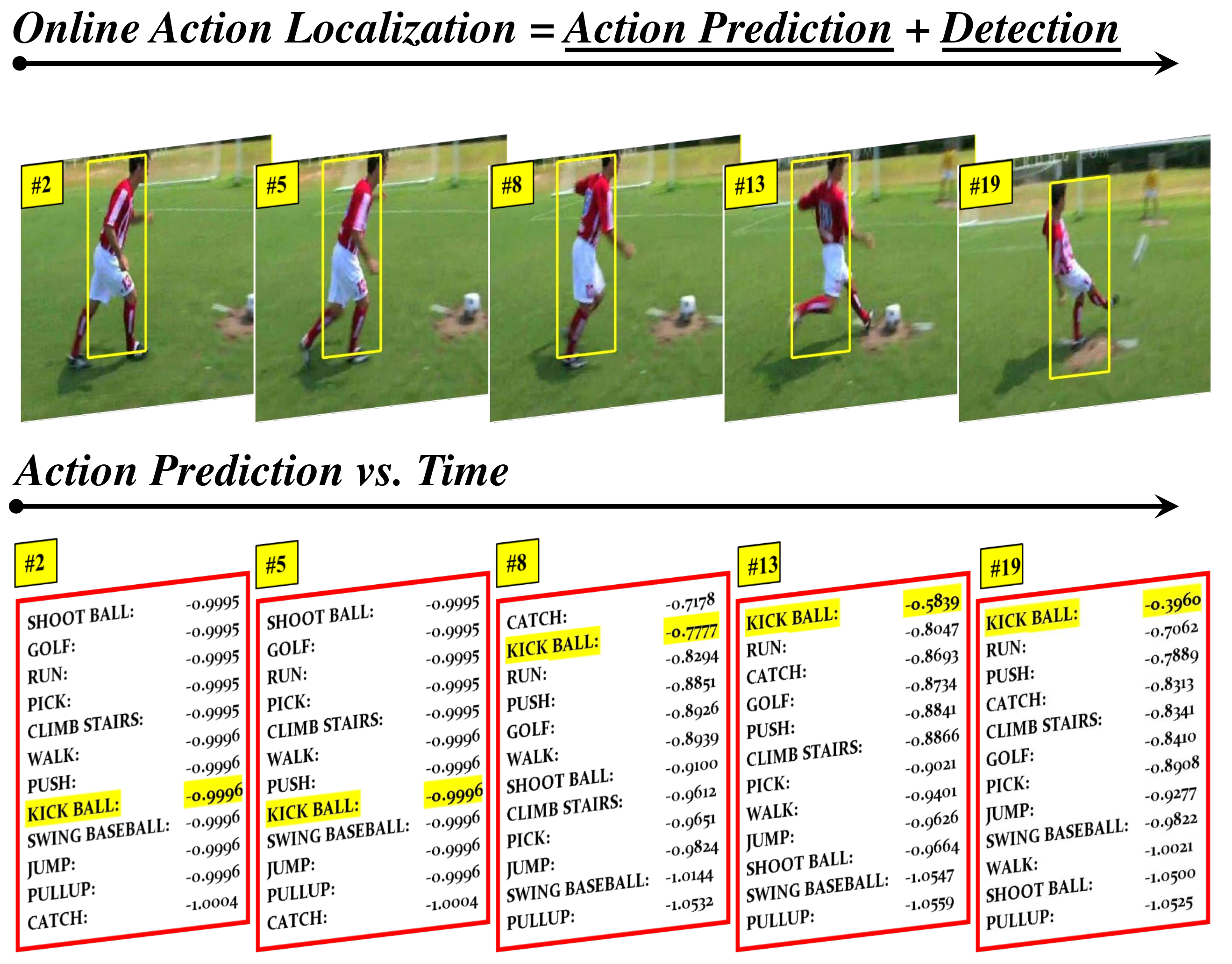}
\caption{This figure illustrates the problem of
\emph{Online Action Localization} that we address in this paper. The top row shows \emph{kick ball} action being performed by a soccer player with frame number shown in top-left of each frame. The goal is to localize the actor (shown with yellow rectangles in top row) and predict the class label of the action (shown in red boxes in second row) as the video is streamed. As can be seen in bottom row, the confidence of \emph{kick ball} action increases and comes to the top as more action is observed across time. This problem contrasts with \emph{offline} action localization where action classification and detection is performed after the action or video clip has been observed in its entirety.}
\label{fig:online_act_loc}
%\vspace{-.1in}
\end{figure}

Predicting \textit{what} and \textit{where} an action or interaction will occur is an important and challenging computer vision problem for automatic video analysis \cite{mori11,wang2014video,yu2011fast,dehghan2014improving}. It involves the use of limited motion information in partially observed videos for frame-by-frame localization and label prediction, and has varied applications in many areas. For human-computer or human-robot interaction, it allows the computer to automatically localize and recognize actions and gestures as they occur, or predict the intention of actors, thereafter creating appropriate responses for them. It is especially relevant to the monitoring of elderly, where  detection of certain actions, e.g. \emph{falling}, must trigger an immediate automated response and alert the care giver or a staff member. Moreover, this allows their interactions with other people to be monitored and quantified for overall well-being. In visual surveillance, online localization and prediction can be used for detecting abnormal actions such as assault or interactions of criminal nature, e.g., drug exchange and alert the human monitors in a timely manner. In automated robot navigation or autonomous driving, the timely detection of human actions in the environment will lead to requisite alteration in path or speed, e.g., a child jumping in front of the car. In this paper, we address the very problem of \textit{Online Action and Interaction Localization}, which aims at localizing actions (interactions) and predicting their class labels in a streaming video (see Fig.~\ref{fig:online_act_loc}).

%In many applications associated with monitoring and security, it is crucial to detect and localize A\&IA in a timely fashion. A particular example is detection and localization of undesirable or malicious occurrences.

In this work, for online action (interaction) localization and prediction, we propose to use the high level structural information using pose in conjunction with a superpixel based discriminative actor foreground model that distinguishes the foreground actor from the background. The superpixel-based model incorporates visual appearance using color and motion features, whereas the pose-based model captures the structural cues through joint locations. Using both the foreground and pose models we generate a confidence map, that is later used to locate the action segments by inferring on a Conditional Random Field in an online manner. Since the appearance of an actor changes due to articulation and camera motion, we retrain foreground model as well as impose spatio-temporal smoothness constraints on poses to maintain representation that is both robust and adaptive. As soon as the human actors are localized at the current frame, we proceed to recognize and predict the label of the action (interaction). There can be multiple approaches to perform online prediction, since the windows over which the visual features are accumulated can be defined in various ways. In \cite{soomro2016predicting}, we used a hybrid of binary SVM and dynamic programming on short intervals to predict the class labels in an online manner. However, this requires multiple classifiers to be trained for each sub-action or segment of an action. In this paper, we present an alternate approach that uses Structural SVM, trained with the objective that the score of the action (interaction) over positive instances should increase as time progresses. Finally, we perform rigorous experiments on four action and two interaction datasets, and introduce measures for consistent evaluation across both actions and interactions. %We also compare with a simple Binary SVM baseline that evaluates the classifier on short chunks of the steaming video.

Existing \emph{offline} action localization methods \cite{mori11,wang2014video,yu2011fast,sdpm13,tubelets14,soomro2015action} classify and localize actions after completely observing an entire video sequence. The goal is to localize an action by finding the volume that encompasses an entire action. Some approaches are based on sliding-windows \cite{sdpm13,oneata2014efficient}, while others segment the video into supervoxels which are merged into action proposals \cite{tubelets14,oneata2014spatio,soomro2015action}. The action proposals from either methods are then labeled using a classifier. Essentially, an action segment is classified \textbf{\textit{after}} the entire action volume has been localized. Similarly, the videos are processed for classification \cite{hoai2014talking,kong2012learning}, retrieval \cite{patron2010high,patron2012structured} or localization \cite{ryoo2009spatio} in an offline manner for the case of interactions. Since offline methods have entire video and action segments at their disposal, they can take advantage of observing entire motion of action instances, and for practical purposes do not provide action detection in a timely manner. Similarly, there have been recent efforts to predict activities by early recognition \cite{li2014prediction, lan2014hierarchical, ryoo2011human, kong2014discriminative}. However, these methods only attempt to predict the label of the action without any localization. Thus, the important question about \textit{where} an action is being performed remains unanswered, which we tackle in this work.

In summary, our contributions in this paper can be summarized as follows: 1) We address the problem of \textit{Online Action and Interaction Localization} in streaming videos, 2) by using high-level pose estimation to learn mid-level superpixel-based foreground models at each time instant. 3) We employ spatio-temporal smoothness constraints on joint locations in human poses to obtain stable and robust action segments in an online manner. 4) The label and confidences for action (interactions) segments are {\em predicted} using Structural SVM trained on partial action clips, which enforces the constraint that the confidence of positive samples increases monotonically over time. Finally, 5) we introduce an evaluation measure to quantify performance of action (interaction) prediction and online localization and perform experiments on six action and interaction datasets with a consistent evaluation framework.

Compared to our CVPR 2016 paper \cite{soomro2016predicting}, we extend our approach to interactions and perform experiments on three additional datasets. Moreover, in contrast to the Binary-SVM and dynamic programming hybrid for online prediction, we employ Structural SVM formulation which  requires one classifier per action and is computationally efficient. Furthermore, we also introduce a unified framework of evaluation for actions and interactions. The rest of the paper is organized as follows. In Sec. \ref{sec:relatedwork} we review literature relevant, whereas Sections \ref{sec:localization} and \ref{sec:prediction} cover the technical details of our approach. We report results in Sec. \ref{sec:experiments} and conclude with suggestions for future work in Sec. \ref{sec:conclusion}.

%%%%%%%%% RELATED WORK %%%%%%%%%
\section{Related Work}\label{sec:relatedwork}

\noindent\textbf{Online Prediction} aims to predict actions from partially observed videos \textit{without} any localization. These methods typically focus on maximum use of temporal, sequential and past information to predict labels and their confidences. Li~and~Fu~\cite{li2014prediction} predict human activities by mining sequence patterns, and modeling causal relationships between them. Zhao~\etal~\cite{zhao2013online} represent the structure of streaming skeletons (poses) by a combination of human-body-part movements and use it to recognize actions in RGB-D. Hoai~and~De~la~Torre~\cite{hoai2014max} simulate the sequential arrival of data while training, and train detectors to recognize incomplete events. Similarly, Lan~\etal~\cite{lan2014hierarchical} propose hierarchical `movemes' to describe human movements and develop a max-margin learning framework for future action prediction.

Ryoo~\cite{ryoo2011human} proposed integral and dynamic bag-of-words for activity prediction, and divide the training and testing videos into small segments and match the segments sequentially. In follow-up work, Ryoo~and~Aggarwal~\cite{ryoo2009spatio} treat interacting people as a group and recognize interactions in continuous videos by computing group motion similarities. Similarly, Kong~\etal~\cite{kong2014discriminative} proposed to model temporal dynamics of human actions by explicitly considering all the history of observed features as well as features in smaller temporal segments. Yu~\etal~\cite{yu2012predicting} predict actions using Spatial-Temporal Implicit Shape Model (STISM), which characterizes the space-time structure of the sparse local features extracted from a video. Cao~\etal~\cite{cao2013recognize} perform action prediction
%by dividing each activity into multiple temporal segments, and
by applying sparse coding to derive the activity likelihood at small temporal segments, and later combine the likelihoods for all the segments. For the case of interactions, Huang~and~Katani~\cite{huang2014action} predict the reaction in a two-person setting by modeling it as an optimal control problem. Recently, there have been works on online temporal detection \cite{de2016online,li2016online} without localization. In contrast to these works, we perform both action prediction and localization in an online manner.

\smallskip
\noindent\textbf{\textit{Offline} Localization} has received significant attention in the past few years, both for actions \cite{xie2011unified,hu2009action,desai2012detecting,jain201515} as well as interactions \cite{rota2015real,kong2014modeling}. For actions, the first category of approaches uses either rectangular tubes or cuboid-based representations. Lan~\etal~\cite{mori11} treated the human position as a latent variable, which is inferred simultaneously while localizing an action. Yuan~\etal~\cite{yuan2011discriminative} used branch-and-bound with dynamic programming, while Zhou~\etal~\cite{zhou2015learning} used a split-and-merge algorithm to obtain action segments that are then classified with LatentSVM \cite{felzenszwalb2010object}. Oneata~\etal~\cite{oneata2014efficient} presented an approximation to Fisher Vectors for tractable action localization. Tran~and~Yuan~\cite{tran2012max} used Structural SVM to localize actions with inference performed using Max-Path search method. Ma~\etal~\cite{ma2013action} automatically discovered spatio-temporal root and part filters, whereas Tian~\etal~\cite{sdpm13} developed Spatio-temporal Deformable Parts Model \cite{felzenszwalb2010object} to detect actions in videos and can handle deformities in parts, both in space and time. Recently, Yu~and~Yuan~\cite{yu2015fast} proposed a method for generating action proposals obtained by detecting tubes with high actionness scores after non-maximal suppression.

The second category uses either superpixels or supervoxels as the base representations \cite{oneata2014spatio,tubelets14}. Jain~\etal~\cite{tubelets14} recently proposed a method that extends selective search approach \cite{uijlings2013selective} to videos. They merge supervoxels using appearance and motion costs and produce multiple layers of segmentation for each video. Gkioxari~and~Malik~\cite{gkioxari2014finding} use selective search \cite{uijlings2013selective} to generate candidate proposals for video frames, whose spatial and motion Convolutional Neural Network (CNN) features are evaluated using SVMs. The per-frame action detections are then linked temporally for localization. There have been few similar recent methods for quantifying actionness \cite{chen2014actionness,yu2015fast}, which yield fewer regions of interest in videos. For interaction recognition in videos, Kong~\etal~\cite{kong2012learning,kong2014interactive} learn high-level descriptions called interactive phrases to express binary semantic motion relationships between interacting people. A hierarchical model is used to encode interactive phrases based on latent SVM framework where interactive phrases are treated as latent variables. Wu~\etal~\cite{wu2013human} also decompose interaction video segments into spatial cells and learn relationship between them. Similar to these methods, our approach can delineate contours of actions and interactions, but with the goal of performing prediction and localization in a streaming fashion.

There are recent works for temporal detection using neural networks with reinforcement learning \cite{yeung2016end} or multi-stage CNNs \cite{shou2016temporal} which only localize the actions temporally. Others use sparse representation to find temporal action proposals \cite{heilbron2016fast}, or statistical language models \cite{richard2016temporal} to temporally localize actions in videos. In contrast, we localize actions both temporally and spatially.

\smallskip
\noindent\textbf{Pose for Recognition.} Low-level motion features, both hand-crafted \cite{wang2013action} and deep learned \cite{wang2015action}, have imparted significant gains to the performance of action recognition and localization algorithms. However, human actions inherently consists of articulation which low-level features cannot model explicitly. The compact and low-dimensional nature of high-level representations such as human poses might make them sensitive and unstable for the task of action localization and recognition. Nonetheless, human pose estimation has been successfully employed for action recognition in several works. For instance, Majiwa~\etal~\cite{maji2011action} implicitly capture poses through `poselet activation vector' and later use them for action recognition in static images. Xu~\etal~\cite{xu2012combining} detect poses through \cite{yang2011articulated} and couple them with independently computed local motion features around the joints for action recognition. Wang~\etal~\cite{wang2013approach} also extended \cite{yang2011articulated} to videos and represented videos in terms of spatio-temporal configurations of joints to perform action recognition. Raptis and Cigal~\cite{raptis2013poselet} recognize and detect interactions from videos by modeling poselets as latent variables in a structural SVM formulation. Joint recognition of action and pose estimation in videos was recently proposed by Xiaohan~\etal~\cite{xiaohan2015joint}. They divide the action into poses and their spatio-temporal parts, and model their inter-relationships through And-Or graphs. Pirsiavash~\etal~\cite{pirsiavash2014assessing} predict quality of sports actions by training a regression model from spatio-temporal pose features, to scores from expert judges. Poses were recently used for \textit{offline} action localization by Wang~\etal~\cite{wang2014video}, who detect actions using a unified approach that discovers action parts using dynamical poselets, and the relations between them.

Similarly, several works model and determine head orientation and upper body pose for recognition and localization of interactions. Patron-Perez~\etal~\cite{patron2010high} developed a per-person descriptor which incorporates head orientation and the local spatio-temporal context around each person to detect interactions. Vahdat~\etal~\cite{vahdat2011discriminative} represented each individual by a set of key poses and formulated spatio-temporal relationships among them in their model. The frame-wise interaction model in Patron-Perez~\etal~\cite{patron2012structured} combines local and global descriptors and incorporates visual attention of people by modeling their head orientations. Although Hoai and Zisserman~\cite{hoai2014talking} do not detect poses per se, they develop a technique to detect different upper body configurations each consisting of multiple parts. In contrast to these methods, we use pose in conjunction with low-level features and mid-level superpixels to predict and localize actions (interactions) in an online manner. Our work is at the cross roads of both online prediction and offline localization, in a unified framework for both actions and interactions operable in partially observed videos.

%%%%%%%%% OVERVIEW OF PROPOSED METHOD %%%%%%%%%
\section{Online Localization of Actions and Interactions} \label{sec:localization}
%\section{Online Localization of Actions and Interactions (A\&IA) in Untrimmed Videos} \label{sec:localization}

%In Sec. \, the confidences of different actions are predicted using dynamic programming on SVM scores. \smallskip
\begin{figure*}[t]
\centering
\includegraphics[width=2\columnwidth]{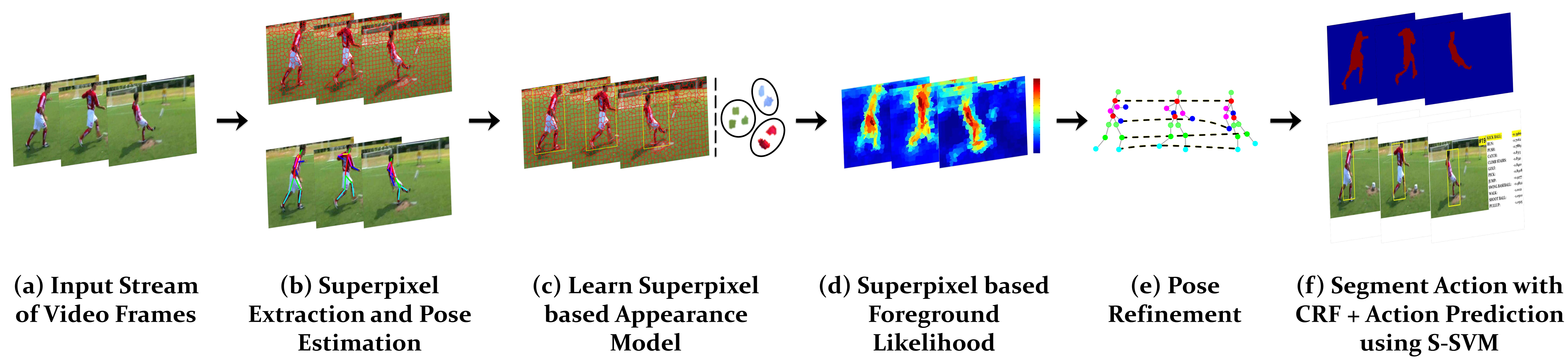}
\caption{This figure shows the framework of the approach proposed in this paper. (a) Given an input video, (b) we over-segment each frame into superpixels and detect poses using an off-the-shelf method \cite{wei2016convolutional}. (c) An appearance model is learned using all the superpixels inside a pose bounding box as positive, and those outside as negative samples. (d) In a new frame, the appearance model is applied on each superpixel of the frame to obtain a foreground likelihood. (e) To handle the issue of visual drift, poses are refined using spatio-temporal smoothness constraints on motion and appearance. (f) Finally, a CRF is used to obtain local action proposals at each frame, on which actions (interactions) are predicted through Structural SVM.}
\label{fig:framework}
%\vspace{-.1in}
\end{figure*}

The pipeline of our approach for localization is shown in Fig. \ref{fig:framework}. Given a testing video, we initialize the localization algorithm with several pose estimations in individual frames and refine the poses using multiple spatio-temporal constraints from previous frames. Next, we segment the testing video frames into superpixels. The features computed within each superpixel are used to learn a superpixel-based appearance model, which distinguishes the foreground from the background by training a discriminative classifier with superpixels within each pose bounding box as foreground and the rest of superpixels as background. Simultaneously, the conditional probability of pose hypotheses at current time-step (frame) is computed using pose confidences and consistency with poses estimated in previous frames. The superpixel and pose-based foreground probability is used to infer the location of actors at each frame through a Conditional Random Field enforcing spatio-temporal smoothness in color, optical flow, motion boundaries and edges among superpixels. After localizing actions (interactions) at each time-step (frame), we refine poses by imposing consistency in locations and appearance of joints as well as scale of poses. Once the pose has been estimated and refined at current time-step, the superpixel-based appearance model is updated to avoid visual drift. This process is repeated for every frame in an online manner (see Fig. \ref{fig:framework}) and gives human localization at every frame. After localization, the spatio-temporal tubes are then used for prediction and recognition of labels at each frame, discussed later in Sec. \ref{sec:prediction}. Thus, the pose estimation not only provides initialization for the proposed discriminative appearance models, as it is more robust compared to human detection in action (interaction) videos due to articulation, it also allows computation of pose features which we use use during label prediction (Sec. \ref{sec:prediction}). Note that the pose estimations can consist of any or multiple body configurations such as upper or full body, as well as multiple humans interacting or performing actions. To simplify the treatment in this section, we assume we are dealing with a single actor or action, without loss of generality.

%Thus, we employ visual and motion cues at multiple levels. Low-level features, such as iDTF \cite{wang2013action} are complemented with high-level structural representations in the form of human poses \cite{desai2012detecting,wei2016convolutional}, and mid-level superpixels which model appearance better than low or high level representations \cite{wang2011superpixel}.

Let $\v s_t$ represent a superpixel by its centroid in frame $t$ and $\v p_t$ represent one of the poses in frame $t$. Since our goal is to localize the actor in each frame, we use $\v X_t$ to represent, a sequence of bounding boxes (tube) in a small window of $\delta$ frames. Each bounding box is represented by its centroid, width and height. Similarly, let $\v S_t$ and $\v P_t$ respectively represent all the superpixels and poses at that time instant. Given the pose and superpixel-based observations till time $t$, $\v S_{1:t}$ and $\v P_{1:t}$, the state estimate $\v X_t$ at time $t$ is obtained using the following equation through Bayes Rule:
%\begin{multline}
% p(\v X_t | \v s_{t-\delta:t}, \v p_{t-\delta:t}) = Z^{-1}p(\v s_t, \v p_t | \v X_t)\\
% \int p(\v X_t | \v X_{t-\delta:t-1}).p( \v X_{t-1} | \v s_{1:t-1}, \v p_{1:t-1} )d \v RX_{t-1},
%\end{multline}
%\begin{multline}
%p(\v X_t | \v s_{t-\delta:t}, \v p_{t-\delta:t}) = Z^{-1}p(\v s_t | \v s_{t-\delta:t}, \v p_{t-1}, \v X_t).\\
%p(\v p_t | \v p_{t-\delta:t-1}, \v X_t). \prod_{\tau=t-\delta}^{t-1} p(\v X_t | \v X_{t-1}),
%\end{multline}
\begin{multline}\label{eq:Bayes}
%\vspace{-.05in}
p(\v X_t | \v S_{1:t}, \v P_{1:t}) = Z^{-1}p(\v S_t | \v X_t).p(\v P_t | \v X_t).\\
\int p(\v X_t | \v X_{t-1}).p( \v X_{t-1} | \v S_{1:t-1}, \v P_{1:t-1} )d \v X_{t-1},
%\vspace{-.05in}
\end{multline}
where $Z$ is the normalization factor, and the state transition model is assumed to be Gaussian distributed, i.e., $p(\v X_t | \v X_{t-1}) = \mathcal{N}(\v X_t; \v X_{t-1}, \v \Sigma)$. Eq. \ref{eq:Bayes} accumulates the evidence over time on the superpixels and poses in streaming mode. The state which maximizes the posterior (MAP) estimate in Eq. \ref{eq:Bayes} is selected as the new state. An implication of Eq. \ref{eq:Bayes} is that the state or localization cannot be altered in the past frames, which makes online localization different from the existing offline methods. Next, we define the pose and superpixel-based foreground likelihoods used for estimating Eq. \ref{eq:Bayes}.

\begin{figure*}
\centering
\includegraphics[width=2.1\columnwidth]{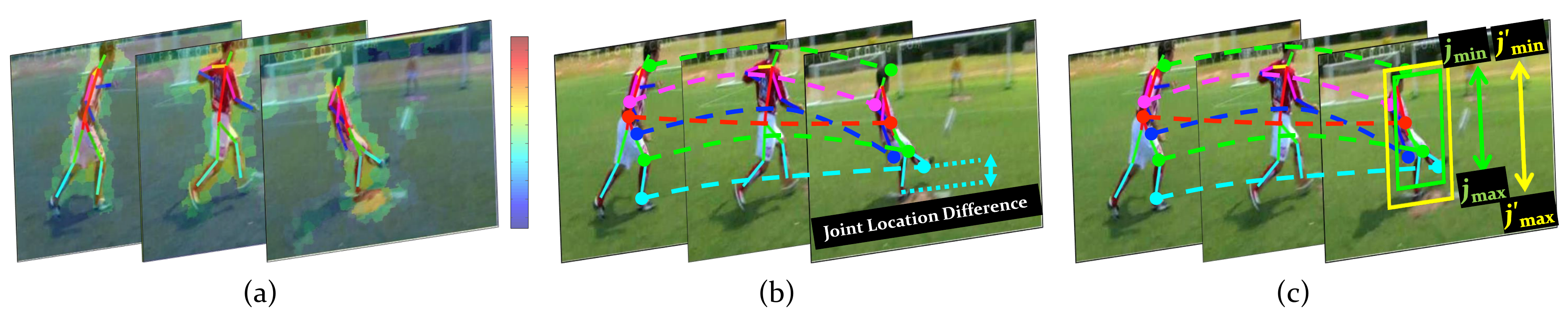}
\caption{This figure shows a visualization of the joint smoothness costs used in pose-based foreground likelihood for (a) appearance smoothness of joints ($J_{\textrm{app}}$), (b) location smoothness of joints ($J_{\textrm{loc}}$) and (c) scale smoothness of joints ($J_{\textrm{sc}}$).}
\label{fig:Joint_Smoothness}
%\vspace{-.1in}
\end{figure*}

\subsection{Superpixel-based Foreground Likelihood}\label{sec:superpixelFG}
Learning an appearance model helps in distinguishing the foreground actions (interactions) from the background. Given foreground and background superpixels in the previous frames $t-\delta, \ldots, t-1$, we group them into $k = 1 \ldots K$ clusters. Furthermore, let $\zeta_k$ define the ratio of foreground to background superpixels for the $k$th cluster through k-means. Then, the appearance-based foreground score using color, $\phi_{\textrm{color}}$, and flow, $\phi_{\textrm{flow}}$, features in the superpixels is given by:
%\vspace{-.05in}
\begin{multline}\label{eq:H_fg}
H_{\textrm{fg}}(\v s_t) = \exp \Big(\frac{\|\bm \phi_{\textrm{color}}(\v s_t) - \v q_k)\|}{r_{k}}\Big) \cdot \zeta_k \\ + \exp \Big(\frac{\|\bm \phi_{\textrm{flow}}\v (s_t) - \bm \mu_k\|}{\rho_k}\Big),
\end{multline}
where $\v q_{k}$ and $r_{k}$ are the cluster center and radius, respectively, whereas $\bm \mu_k$ and $\rho_k$ represent the mean and variance of optical flow for the $k$th cluster.

In Eq. \ref{eq:H_fg}, the clusters are updated incrementally at each time-step (frame) to recover from the visual drift using a temporal window of past $\delta$ frames. Note that, background superpixels within a foreground bounding box are inevitably considered as foreground initially, however the later segmentation through Conditional Random Field serves to alleviate this problem by separating foreground superpixels within the bounding box localization. The $\zeta_k$ helps to compensate for this issue by quantifying the foreground/background ratio for each cluster. Finally, the superpixel-based foreground likelihood in Eq. \ref{eq:Bayes} is given as: $p(\v S_t | \v X_t) = \alpha_{\textrm{fg}} \cdot H_{\textrm{fg}}(\v s_t)$, where $\alpha_{\textrm{fg}}$ is the normalization factor.

%\begin{multline}\label{eq:H_app}
%H_{\textrm{app}}(\v s_t, \v p_{t-1}) = Z^{-1} \bigg( \exp \Big(\frac{\|\v \phi_{\textrm{col}}(\v s_t) - \v c_k)\|}{r_{k}}\Big) \cdot \xi_k + \\
%\sum_{j=1}^{|\prescript{*}{}{\Pi}_{t-1}|} \exp\big( - \| \v s_{t} - \prescript{*}{}{\bm \pi}_{t-1} \| \big) \bigg),
%\end{multline}

\subsection{Pose-based Foreground Likelihood}
\label{sec:poseFG}
We represent each pose $\v p_t$ graphically with a tree, given by $\v T = (\v \Pi,\v \Lambda)$. The body joints $\v \pi \in \v \Pi$ are based on appearance connected by $\v \lambda \in \v \Lambda$ edges capturing the structure of the pose. The joint $j$ with its location in pose $\v p_t$ is represented by ${\bm \pi}^j_t$, consisting of its $x$ and $y$ locations. %= [x^j_t,y^j]$. %The location of a joint $j$ for pose $\v p_t$ is given by $({x}^j_t, {y}^j_t)$.
Then, the raw cost (or negated detection score) for a particular pose $\v p_t$ is the sum of appearance and deformation costs:
%\vspace{-.05in}
\begin{equation}\label{eq:H_raw}
H_{\textrm{raw}}(\v p_t) = \sum_{j \in \v \Pi_t}^{}{\v \psi\big({\bm \pi}^j_t\big)} + \sum_{(j,j') \in \v \Lambda_t}{\v \chi \big( {\bm \pi}^j_t, {\bm \pi}^{j'}_t \big)},
\end{equation}
where $\v \psi$ and $\v \chi$ are linear functions of appearance features of pose joints, and the relative joint displacements (deformations) w.r.t each other. We use a pre-trained pose detector to obtain pose hypotheses in each frame. In \cite{soomro2016predicting}, we used Flexible Mixture-of-Parts \cite{yang2011articulated} for pose estimation, which optimizes over latent variables that capture different joint locations and pose configurations. In this paper, we report results using Convolutional Pose Machines (CPM) \cite{wei2016convolutional} which uses deep learning. For CPM, the deformation costs are embedded within joint costs in Eq. \ref{eq:H_raw}. Since the pose estimation in both methods works on individual frames, it is inherently noisy and does not take into account the temporal information available in videos. We impose the following smoothness constraints (as shown in Fig. \ref{fig:Joint_Smoothness} (a-c)) in the previous $\delta$ frames to re-evaluate poses in Eq. \ref{eq:H_raw} for the current time-step.

\medskip
\noindent\textbf{\textit{Appearance Smoothness of Joints:}} Since the appearance of a joint is not expected to change drastically in a short window of time, we impose the appearance consistency between superpixels at joint locations:
%\vspace{-.05in}
\begin{equation}\label{eq:J_app}
J_{\textrm{app}} (\v p_t) = \sum_{j=1}^{|\v \Pi_t|} \| H_{\textrm{fg}}(\hat{\v s}^j_t) - H_{\textrm{fg}}(\hat{\v s}^j_{t-1})\|,
%\vspace{-.1in}
\end{equation}
where $\hat{\v s}^j_t$ is the enclosing superpixel of the joint ${\bm \pi}^j_t$.

%\begin{equation}\label{eq:J_app}
%J_{\textrm{app}} (\v p_t) = \sum_{j=1}^{|\Pi_t|} \| H_{\textrm{fg}}(\hat{\v s}_t) - H_{\textrm{fg}}(\hat{\v s}_{t-1})\|,
%\end{equation}
%where $\hat{\v s}_t$ is the overlapping superpixel of the joint ${\bm \pi}^j_t$.

\medskip
\noindent\textbf{\textit{Location Smoothness of Joints:}} Since human motion is naturally smooth, we ensure that displacements in joint locations over time are small. This is achieved by fitting a 2D spline using piecewise polynomials to each joint $j$ on the past $\delta$ frames, ${\bm \gamma}^j_{t}$. Then the location smoothness cost over all joints is given by:
%\vspace{-.05in}
\begin{equation}\label{eq:J_loc}
J_{\textrm{loc}}(\v p_{t}) = \sum_{j=1}^{|\v \Pi_t|} \| {\bm \gamma}^j_{t} - {\bm \pi}^j_{t} \|.
%\vspace{-.1in}
\end{equation}

\medskip
\noindent\textbf{\textit{Scale Smoothness of Joints:}} Let $j_{\textrm{min}}$ and $j_{\textrm{max}}$ denote the vertical minimum and maximum for all the splines $\bm \gamma_\tau, \forall \tau \in \{t-\delta, \ldots, t\} $, i.e., the $y$-axis components of the bounding box circumscribing all the splines fitted on joints. Furthermore, let $j'_{\textrm{min}}, j'_{\textrm{max}}$ denote minimum and maximum for joints in actual poses ${\bm \pi}_t \in \v \Pi_t$. Then, the scale smoothness cost essentially computes the overlap between the two heights:
%\vspace{-.05in}
\begin{equation}\label{eq:J_sc}
J_{\textrm{sc}}(\v p_t) = \| (j_{\textrm{max}} - j_{\textrm{min}}) - (j'_{\textrm{max}} - j'_{\textrm{min}}) \|.
\end{equation}

\medskip
The combined cost of a particular pose is defined as its raw cost plus the smoothness costs across space and time, i.e.,
\begin{equation}\label{eq:H_pose}
H_{\textrm{pose}}(\v p_t) = H_{\textrm{raw}}(\v p_t) + J_{\textrm{app}}(\v p_t) + J_{\textrm{loc}}(\v p_t) + J_{\textrm{sc}}(\v p_t).
\end{equation}
The change in pose and appearance of an actor may cause visual drift. Similar to Sec. \ref{sec:superpixelFG}, we use a temporal window of past $\delta$ frames to refine the pose locations. This helps in better prediction of the highly probable foreground locations in current frame.
We propose an iterative approach to select poses in the past $\{t-\delta, \ldots, t\}$ frames. Given an initial set of poses, we fit a spline to each joint ${\bm \pi}^j_{t}$. Then, our goal is to select a set of poses from $t-\delta$ to $t$ frames, such that the following cost function is minimized:
\begin{multline}\label{eq:object}
%\vspace{-.1in}
%\hspace{-.11in}
(\prescript{*}{}{\v p}_{t-\delta}, \dots, \prescript{*}{}{\v p}_{t}) = \argmin_{\v p_{t-\delta}, \dots, \v p_t} \sum_{\tau = t-\delta}^{t} \bigg(H_{\textrm{pose}}(\v p_\tau) \bigg).%\hspace{-.11in}
%\vspace{-.1in}
\end{multline}

This function optimizes over pose detection, and the appearance, location and scale smoothness costs of joints (see Fig.\ref{fig:framework} (e)) by greedily selecting the minimum cost pose in every frame through multiple iterations, such that the joints are spatially accurate and temporally consistent with the motion of the action. This procedure is summarized in Algorithm \ref{alg:poseRefine}. Note that the poses in previous frames of the batch are only refined simultaneously, however, the pose at current time step is used by the algorithm. Finally, the pose-based foreground likelihood in Eq. \ref{eq:Bayes} is given by $p(\v P_t | \v X_t) = \exp(\alpha_{\textrm{pose}} \cdot H_{\textrm{pose}}(\v p_t))$, where $\alpha_{\textrm{pose}}$ is the normalization factor.

\begin{algorithm}
\caption{: Algorithm to refine pose locations in a batch of frames in $Q$ iterations.}
\label{alg:poseRefine}

\textbf{Input}: $\v P_{t-\delta}, \ldots, \v P_{t}$ \\
%and $\v S_{t-\delta}, \ldots, \v S_{t}$  \\
\textbf{Output}: $\prescript{*}{}{\v p}_{t-\delta}, \dots, \prescript{*}{}{\v p}_t$

\rule{1\linewidth}{0.02cm}

\begin{algorithmic}[1]
\Procedure{RefinePoses}{$ $}
	\For{$\tau = t-\delta$ to $t$}
		\State  $\prescript{*}{}{\v p}_{\tau} = \argmin ( H_{\textrm{raw}}(\v p_{\tau}))$
    \EndFor

	\For{$n = 1$ to $Q$}
		\State Fit a spline $\bm \gamma^j$ to each joint using locations \\ \hspace{.45in}$[\prescript{*}{}{\bm \pi}^j_{t-\delta},\ldots,\prescript{*}{}{\bm \pi}^j_{t}]$	
        \State Compute $J_{\textrm{app}}({\v p}_{t})$ using Eq. \ref{eq:J_app}
	    \State Compute $J_{\textrm{loc}}({\v p}_{t})$ using Eq. \ref{eq:J_loc}
       \State Compute $J_{\textrm{sc}} (\v p_t)$ using Eq. \ref{eq:J_sc}
        \State Find $(\prescript{*}{}{\v p}_{t-\delta}, \dots, \prescript{*}{}{\v p}_t)$ through Eq. \ref{eq:object}.
	\EndFor
		
\EndProcedure
\end{algorithmic}
\end{algorithm}

\subsection{Actor Segmentation using Conditional Random Fields (CRF)}

Once we have the superpixel and pose-based foreground likelihoods in Eq. \ref{eq:Bayes}, we proceed to infer the action segment and its contour using a history of $\delta$ frames. Although the action location is computed online for every frame, using past $\delta$ frames adds robustness to segmentation. We form a graph
%$\v G(\v V,\v E)$
with superpixels as nodes connected through \textit{spatial} and \textit{temporal} edges. %Spatial edges are formed between neighboring superpixels in a single frame, whereas temporal edges are formed between overlapping superpixels and their neighbors in adjacent frames.
Let variable $a$ denote the foreground/background label of a superpixel. Then, the objective function of CRF becomes:
\begin{multline}\label{eq:MRF}
%\vspace{-.05in}
-\log\big(p( a_{t-\delta},\dots,a_t| \v s_{t-\delta},\ldots,\v s_{t}, \v p_{t-\delta},\ldots,\v p_{t}) \big) \\= \sum_{\tau=t-\delta}^{t} \Big( \underbrace{\v \Theta \big( a_{\tau} | \v s_{\tau}, \v p_{\tau} \big)}_{\textrm{unary potential}} + \underbrace{\v \Upsilon \big(a_{\tau},a'_{\tau} | \v s_{\tau}, \v s'_{\tau} \big)}_{\textrm{spatial smoothness}} \Big) \\
+ \sum_{\tau=t-\delta}^{t-1} \underbrace{\v \Gamma \big(a_{\tau},a'_{\tau+1} | \v s_{\tau}, \v s'_{\tau+1} \big)}_{\textrm{temporal smoothness}},
%\vspace{-.05in}
\end{multline}
%\big( a^i | \v v^i, \v \Psi^T \big)\\
%&\hspace{.55in}+ \sum_{\v v^j | \v e^{ij} \in \v E} \Upsilon \big(a^i,a^j | \v v^i, \v v^j, \v \Phi^i, \v \Phi^j; w_\Upsilon \big) \Big),
where the unary potential, with the associated weights symbolized with $\alpha$, is given by:
\begin{multline}\label{eq:MRF_unary}
\v \Theta \big( a_{\tau} | \v s_{\tau}, \v p_{\tau} \big) = \alpha_{\textrm{fg}}H_{\textrm{fg}}(\v s_{\tau}) + \alpha_{\textrm{pose}}H_{\textrm{pose}}(\v p_{\tau}),
\end{multline}
%where the unary potential $\Theta \big( a_{\tau} | \v s_{\tau}, \v p_{\tau} \big) = H_{\textrm{fg}}(\v s_{\tau}) + H_{\textrm{pose}}(\v p_{\tau})$,
and the spatial and temporal binary potentials, with weights $\beta$ and distance functions $d$, are given by:
\begin{multline}\label{eq:MRF_spatial_binary}
\v \Upsilon \big(a_{\tau},a'_{\tau} | \v s_{\tau}, \v s'_{\tau} \big) \\ = \beta_{\textrm{col}}d_{\textrm{col}}(\v s_{\tau}, \v s'_{\tau}) + \beta_{\textrm{hof}}d_{\textrm{hof}}(\v s_{\tau}, \v s'_{\tau}) + \beta_{\mu}d_{\mu}(\v s_{\tau}, \v s'_{\tau}) \\+ \beta_{\textrm{mb}}d_{\textrm{mb}}(\v s_{\tau}, \v s'_{\tau}) + \beta_{\textrm{edge}}d_{\textrm{edge}}(\v s_{\tau}, \v s'_{\tau}),
%\vspace{-.05in}
\end{multline}
and
\begin{multline}\label{eq:MRF_temporal_binary}
%\vspace{-.05in}
\v \Gamma \big(a_{\tau},a'_{\tau-1} | \v s_{\tau}, \v s'_{\tau-1} \big) = \beta_{\textrm{col}}d_{\textrm{col}}(\v s_{\tau}, \v s'_{\tau-1}) \\+ \beta_{\textrm{hof}}d_{\textrm{hof}}(\v s_{\tau}, \v s'_{\tau-1}) + \beta_{\mu}d_{\mu}(\v s_{\tau}, \v s'_{\tau-1}),
\end{multline}
respectively. In Eqs. \ref{eq:MRF_spatial_binary}, and \ref{eq:MRF_temporal_binary}, $\beta_{\textrm{col}}d_{\textrm{col}}(.)$ is the cost of color features in HSI color space, $\beta_{\textrm{hof}}d_{\textrm{hof}}(.)$ and $\beta_{\mu}d_{\mu}(.)$ compute compatibility between histogram of optical flow and mean of optical flow magnitude of the two superpixels, respectively. Similarly, $\beta_{\textrm{mb}}d_{\textrm{mb}}(.)$ and $\beta_{\textrm{edge}}d_{\textrm{edge}}(.)$ quantify incompatibility between superpixels with prominent boundaries.

\section{Online Prediction of Actions and Interactions}\label{sec:prediction}
%\section{Online Prediction of Actions and Interactions in Untrimmed Videos}\label{sec:prediction}

%After obtaining the localized segments at each time instant, we proceed to predict the action label within the localized segment.
% box through dynamic programming using scores from Support Vector Machines (SVMs) on short video clips. These SVMs were trained on temporal segments of the training videos.

For online recognition and class-label prediction of actions (interactions) in streaming videos, the classifier has to be applied on-the-fly on short temporal intervals. In particular, training videos of an action (interaction) class $c$ are divided into $M$ clips of equal-sized interval $\Omega$. The average length of each segment is saved as prior information, which during testing allows us to compute features in intervals of the desired length. Next, we present a baseline approach using Support Vector Machine and Dynamic Programming hybrid (from our CVPR 2016 paper \cite{soomro2016predicting}) which divides videos into short segments, and trains a classifier independently for each segment. The online update of action confidences is achieved through dynamic programming on segment scores. In this paper, we present an alternate approach which makes structured prediction by training a single classifier per action and modeling temporal dependence between action segments. In this section, we present the formulation in terms of linear classifiers for simplicity, however, in practice we used SVM with histogram intersection kernel.

Let $m$ index over temporal segments, i.e., $m \in {1, \ldots, M}$ and $\v x_{i,m}$ denote the $m$th segment as well as its feature vector in video $i$. Next, we present the two approaches we use to recognize and predict the class label at time step $t$ of a testing video.

%\subsection{Binary SVM}

%with a separate Structural SVM trained on each time interval, i.e., $0 \rightarrow 1 \;\textrm{sec}, 1 \rightarrow 2 \;\textrm{sec}, \ldots$ from all the videos of a particular action.

\subsection{Binary SVMs with Dynamic Programming Inference (DP-SVM)} \label{subsec:SVM_DP}

First, we present a baseline for online prediction \cite{soomro2016predicting} in our localization framework. For training binary SVMs for segments in an action (interaction) class $c$, we assume availability of $N$ trimmed positive and negative training videos. For linear SVM, we obtain a single weight vector $\v w_m$ per segment by optimizing the following objective function,
\begin{align}\label{eq:binarySVM}
\textrm{min} & \hspace{.2in } \frac{1}{2} \| \v w_m\|^2 + C \sum_{i=1}^N\sum_{m=1}^{M} \xi_{i,m} \nonumber\\
\textrm{s.t.}& \hspace{.2in } \v y_{i,m}\langle \v w_m,\v x_{i,m} \rangle \geq 1 - \xi_{i,m}, \; \xi_{i,m} \geq 0, & \forall i,m
\end{align}
where $C$ controls the trade-off between regularizer and constraints, and $\v y_{i,m} = 1$, for desired $m$ if $i \in c$ and $-1$, otherwise. Effectively, the training videos are divided into short intervals and an SVM is trained for each interval $m$ independently. While testing on videos, the classification is performed on features accumulated on interval lengths learned from training videos. To exploit and preserve the sequential information present in videos, this is followed by dynamic programming on the short interval clips. At each step of the dynamic programming, the system effectively searches for the best matching segment that maximizes the SVM confidences from past segments. This method is applied independently for each class, and gives the confidence for that class. This shares resemblance to Dynamic Bag-Of-Words~\cite{ryoo2011human} which used RBF function to compute score between training and testing segments, and applied it on trimmed videos.

Let $F(t, z)$ be the result of dynamic programming at time $t$ assuming the current interval is $z$ for a particular class. The result of applying classifier on testing video $o$ on features computed between $t-\Omega$ and $t$ is given by $\sigma(\langle \v w_m , \v x_{o,t} \rangle)$, where $\sigma$ is the sigmoid function. If the testing video is trimmed, then $F(t,z)$ is computed using the following recursion:
\begin{equation}
F(t,z) = \max_{m}F(t-\Omega, z-m) \cdot \sigma(\langle \v w_m , \v x_{o,t} \rangle).
\end{equation}

At each time instant, the maximum value at time $t$ gives the desired confidence for the class under consideration.
%(untrimmed version)

\subsection{Structural SVM (S-SVM)}
\label{subsec:SSVM}

Ideally the prediction confidence for the correct class should increase as more action (interaction) in the video is observed over time. There is rich structure that can be derived from division of actions into sub-actions, and modeling the spatio-temporal dependence between them.%, as well as presence of multiple actors during interactions.
Given testing video segments, we then apply Structural SVM detector to each segment of the test video. For this case, we redefine intervals w.r.t start time of an action (interaction), i.e., the start time of interval $m$ is $0$ of the trimmed training video. We set the problem in a Structural Support Vector Machine (S-SVM) with margin re-scaling construction, given by:

\begin{align}\label{eq:structuralSVM}
%\vspace{-.05in}
\textrm{min} & \hspace{.2in } \frac{1}{2} \| \v w\|^2 + C \sum_{i=1}^{N*M} \xi_i \nonumber\\
\textrm{s.t.}& \hspace{.2in } \langle \v w,\v \Psi_i(\v x_i, \v y_i) - \v \Psi_i(\v x_i, \v y) \rangle \geq \v \Delta(\v y_i, \v y) - \xi_i, \nonumber\\
& \hspace{.2in }\forall \v y \in \mathcal{Y}\; \backslash \; \v y_i, \xi_i \geq 0,  \forall i
\end{align}
where the joint feature map for input and output is given by $\v \Psi(\v x, \v y) = \v x \cdot \textrm{sign}(\v y)$, and $\mathcal{Y} = \{-1, 1, \ldots, M\}$ is the set of all labels . In Eq. \ref{eq:structuralSVM}, $\xi$ represents the slack variables for the soft-margin SVM, which optimizes over the learned weight vector $\v w$ and the slack variables $\xi$. The constraint with the loss function $\v \Delta(\v y_i, \v y)$ ensures that the score with the correct label $\v y_i$ is greater than alternate labels. Since the number of constraints can be tremendous, only subset of constraints are used during the optimization. For each training sample, the label $\v y$ which maximizes $\langle \v w, \v \Psi(\v x_i, \v y)\rangle + \v \Delta(\v y_i, \v y)$ is found and the constraint which maximizes this loss is added into the subset, known as the most violated constraint. For both actions and interactions, the temporal component of the loss is defined as:
\begin{equation}\label{eq:loss_time}
    \v \Delta(\v y_i, \v y_{i'})=
\begin{cases}
    |\v y_i - \v y_{i'}|, & i \in c \wedge i' \in c\\
    M + \epsilon,    & i \in c \wedge i' \notin c\\
    \epsilon,        & \text{otherwise}.
\end{cases}
\end{equation}
%Furthermore, for individual labels in both actions and interactions, we use a $0-1$ loss, whereas
The above loss function ensures that the confidence increases as the action (interaction) happens in the testing video, i.e., the evaluation during a positive test instance, possibly over a long video, yields a unique signature of confidence values that increases over time. This approach results in one S-SVM per class, and can be applied indiscriminately to untrimmed videos. For interactions an additional loss captures the relationship between actors. Once the weight vector $\v w$ has been learned, the score for a clip in the testing video is computed using $\argmax_{\v y \in \mathcal{Y}} \langle \v w, \v \Psi (\v x, \v y) \rangle$.

Note that the performance of detection or prediction for action (interaction) localization depends on the quality of localized tubes / cuboids, as the classifiers are only evaluated on such video segments. This is in contrast to previous prediction methods \cite{ryoo2011human,li2014prediction,kong2014discriminative,hoai2014max} which do not spatially localize the actions (interactions).

%%%%%%%%% EXPERIMENTS %%%%%%%%%
\section{Experiments}\label{sec:experiments}

%\begin{figure*}
%\centering
%\includegraphics[width=2.1\columnwidth]{./figAUC_Time.pdf}
%\caption{This figure shows action prediction and localization performance as a function of observed video percentage. (a) shows prediction accuracy for JHMDB and UCF Sports datasets; (b) and (c) show localization accuracy for JHMDB and UCF Sports, respectively. Different curves show evaluations at different overlap thresholds: $10\%$ (red), $30\%$ (green) and $60\%$ (pink).}
%\label{fig:finalOnlineAUC}
%%\vspace{-.1in}
%\end{figure*}

We evaluate our \textit{online action localization} approach on six challenging datasets: 1) JHMDB, 2) Sub-JHMDB, 3) MSR-II, 4) UCF Sports, 5) TV Human Interaction and 6) UT Interaction datasets. We provide details for the experimental setup followed by the performance evaluation and analysis of the proposed approach.

\smallskip
\noindent\textbf{Features:} For each frame of the testing video we extract superpixels using SLIC \cite{achanta2012slic}. This is followed by extraction of color features (HSI) for each superpixel in the frame, as well as improved Dense Trajectory features (iDTF: HOG, HOF, MBH, Traj) \cite{wang2013action} within the streamed volumes of the video. Each superpixel descriptor has a length of $512$ and we set $K = 20$. The pose detections are obtained using \cite{wei2016convolutional} and pose features using \cite{jhuang2013towards}.
%, which consist of $9$ features (joint locations: $4$, joint trajectories: $5$). The location features capture spatial relationships between joints, while trajectory features capture motion of joints over time.
We build a vocabulary of $20$ words for each pose feature, and represent a pose with $180$d vector.

\smallskip
\noindent\textbf{Parameters and Distance Functions:} We use Euclidean distance for $d_{\mu}$, chi-squared distance for $d_{\textrm{hof}}$ and $d_{\textrm{col}}$, and geodesic distance for $d_{\textrm{mb}}$ and $d_{\textrm{edge}}$. We normalize the scores used in CRF, therefore, we set absolute values of all the parameters $\alpha$ and $\beta$ to 1.

\smallskip
\noindent\textbf{Evaluation Metrics:} Since the online localization algorithm generates tubes or cuboids with associated confidences, the Receiver Operating Characteristic (ROC) curves are computed at fixed overlap thresholds. Following experimental setup of \cite{mori11}, we show ROC @ $20\%$ overlap. Furthermore, Area Under the Curve (AUC) of ROC at various thresholds gives an overall measure of performance. The proposed evaluation metrics are computed over all action and interaction datasets for consistency. For MSR-II dataset, we also report results using Precision and Recall curves typically used for this dataset.

Inspired from early action recognition and prediction works \cite{ryoo2011human}, we also quantify the performance as a function of \textit{Observation Percentage} of actions (interactions). For this evaluation method, the localization and classification for testing videos are sampled at different percentages of observed video/action ($0, 0.1,0.2,\ldots,1$). The ROC curve is computed at multiple overlap thresholds, and AUC is computed under ROC curves at respective thresholds. In the case of untrimmed videos, we evaluate the prediction accuracy as a function of observation percentage within the temporal boundaries of actions (interactions).

Note that, in online action (interaction) localization, the prediction and localization is performed instantaneously at each frame in a streaming video, therefore once locations are detected and predictions are made, retroactive modifications or changes to results are not possible. %This leads to finding the spatial overlap with the ground truth, instead of spatio-temporal.

%\smallskip
%\noindent\textbf{Baseline for Online Action Localization:} We compare with offline methods which use entire videos to localize actions, and also compute results for a competitive online localization baseline for comparison. For the proposed baseline, we exhaustively generate bounding boxes with overlaps at multiple scales in each frame. These boxes are connected with appearance similarity costs. Over time, the boxes begin to merge into tubes. For temporal window of $\delta=5$ frames (same as our method), we evaluate each tube with classifiers for all the actions using iDT features.

\begin{figure*}
\centering
\includegraphics[width=1.6\columnwidth]{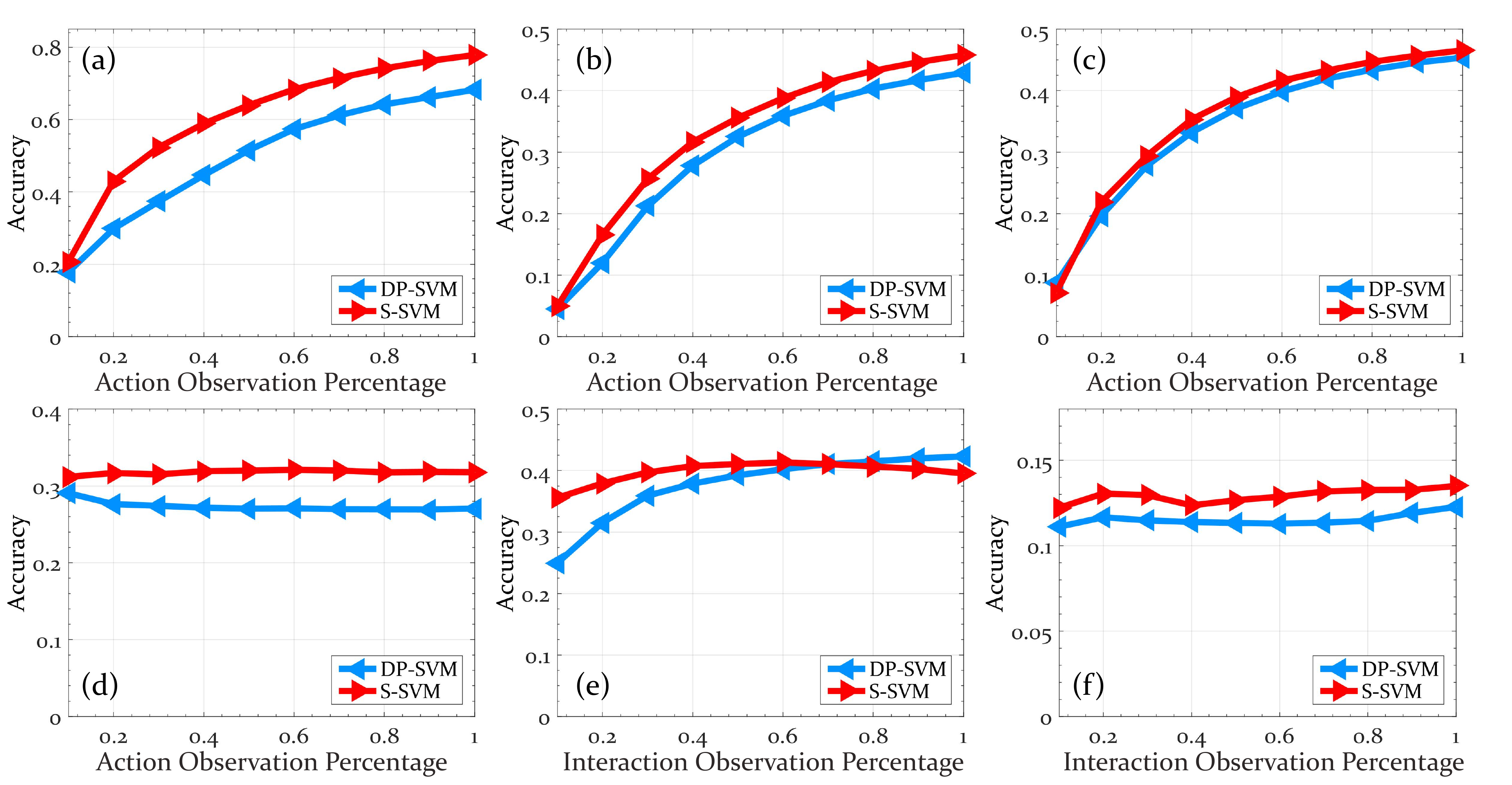}
\caption{This figure shows action prediction performance as a function of observed percentage of action or interaction for (a) UCF Sports, (b) JHMDB, (c) sub-JHMDB, (d) MSR-II, (e) TV Human Interaction and (f) UT Interaction datasets. Prediction performance by the baseline Binary SVM with Dynamic Programming approach is shown in blue, and that of Structural SVM with the red curve.}
\label{fig:PredvsTime}
%\vspace{-.1in}
\end{figure*}

\subsection{Datasets}
%\smallskip
\noindent\textbf{JHMDB Dataset:} The JHMDB \cite{jhuang2013towards} dataset is a subset of the larger HMDB51 \cite{kuehne2011hmdb} dataset collected from digitized movies and YouTube videos. It contains $928$ videos consisting of $21$ action classes. The dataset has annotations for all the body joints and has recently been used for offline action localization \cite{gkioxari2014finding}. We use a codebook size of $K=4000$ to train SVMs using iDTF features.

%Fig. \ref{fig:final_subJHMDB_JHMDB}(c) shows the results of the proposed method in red, and that of \cite{gkioxari2014finding} in blue. The difference in performance is attributed to the online vs. offline nature of methods, as well as the use of CNN features by \cite{gkioxari2014finding}. We also provide a competitive online localization baseline (in gray) and in comparison we obtain better performance.

\smallskip
\noindent\textbf{sub-JHMDB Dataset:} The sub-JHMDB dataset has all human body joints visible in each frame. It contains a total of $316$ videos over $12$ action classes: \textit{catch, climb stairs, golf, kick ball,} etc. The presence of the entire human within each frame makes it more challenging to recognize and localize the actions as compared to JHMDB dataset, due to high articulation of human body joints and complex variations in appearance and motion. A codebook size of $K = 4000$ was used for IDTF, and SVMs were trained with a bag-of-words representation inside the ground truth action volumes.

%We report quantitative results on this dataset using ROC at $20\%$ overlap in Fig. \ref{fig:final_subJHMDB_JHMDB}(a) and AUC in Fig. \ref{fig:final_subJHMDB_JHMDB}(b). The graphs show that we get competitive results compared to the state-of-the-art method of Wang \etal \cite{wang2014video}, who also evaluated a competitive baseline over this dataset, which uses iDT features with a Fisher Vector encoding (black curves in Fig. \ref{fig:final_subJHMDB_JHMDB}(a,b)). For the baseline, they exhaustively scan at various spatio-temporal locations at multiple scales in the video. Despite being an online method using BoW representation, the proposed approach clearly outperforms the IDTF + FV baseline, and emphasizes the strengths of our approach.

\smallskip
\noindent\textbf{UCF-Sports Dataset:} The UCF Sports \cite{mikel2008,soomro2014action} dataset is collected from broadcast television channels and consists of $150$ videos. It includes a total of $10$ action classes: \textit{diving, golf swing, kicking, lifting, riding horse, skateboarding,} etc. Videos are captured in a realistic setting with intra-class variations, camera motion, background clutter, scale and viewpoint changes. We evaluated our method using the methodology proposed by \cite{mori11}, who use a train-test split with intersection-over-union criterion at an overlap of $20\%$.
To train SVM, we use a codebook size of $K = 1000$ on iDTFs using all the training videos.

%We show a quantitative comparison with existing \textit{offline} state-of-the-art methods using ROC @ $20\%$ and Area Under Curve (AUC) in Fig. \ref{fig:AUC_UCFSports}.

%\smallskip
%\noindent\textbf{UCF Sports Dataset:} The UCF Sports \cite{mikel2008,soomro2014action} dataset consists of $150$ videos with $10$ action classes. We evaluated our approach using the methodology proposed by \citet{mori11}, with a train-test split and intersection-over-union criterion at an overlap of $20\%$.
%To train SVMs, we use a codebook size of $1000$ on iDTFs. % using all the training videos.

\smallskip
\noindent\textbf{MSR-II Dataset:} The MSR-II dataset \cite{yuan2011discriminative} consists of $54$ untrimmed videos and $3$ action classes: Boxing, Handclapping and Handwaving. We follow the experimental methodology of \cite{yuan2011discriminative}, having cross-dataset evaluation, where KTH \cite{schuldt2004recognizing} dataset is used for training and testing is performed on MSR-II dataset. A codebook size of $K = 1000$ was used to train SVM on iDTFs. We show quantitative comparison using Precision-Recall curves with state-of-the-art \textit{offline} methods. However, for uniformity with other datasets we also report results using ROC and AUC curves.

%in Fig. \ref{fig:Fig_MSR2_Small}.

\smallskip
\noindent\textbf{TV Human Interaction (TVHI):} The TVHI dataset \cite{patron2010high,patron2012structured} is collected from $23$ different TV shows and is composed of $300$ untrimmed videos. It includes $4$ interaction classes: \textit{hand shake, high five, hug and kiss}, with $50$ videos each. It also contains a negative class with $100$ videos, that have none of the listed interactions. The videos have varying number of actors in each scene, different scales and abrupt changes in camera viewpoint at shot boundaries. For our experiments we only use the $4$ interaction classes (excluding negative class) for interaction localization. We use the suggested experimental setup of two train/test splits. The localization performance is reported using ROC and AUC curves.

\smallskip
\noindent\textbf{UT Interaction:} The UT Interaction dataset \cite{UT-Interaction-Data,ryoo2009spatio} contains untrimmed videos of $6$ interaction classes: \textit{hand-shaking, hugging, kicking, pointing, punching, and pushing.} Similar to \cite{patron2012structured}, we add \textit{being kicked, being punched} and \textit{being pushed} as interactions. The dataset consists of two sets, where each set has $10$ video sequences and each sequence having at least one execution per interaction. Videos involve camera jitter with varying background, scale and illumination. We follow the recommended experimental setup by using $10$-fold leave-one-out cross validation per set. That is, within each set we leave one sequence for testing and use remaining $9$ for training. We report the average localization performance of the proposed approach using ROC @ $20\%$ overlap and AUC curves. %For our experiments include two-person interaction classes and exclude the \textit{pointing} interaction class, which is based on a single person.

\begin{figure*}
\centering
\includegraphics[width=1.7\columnwidth]{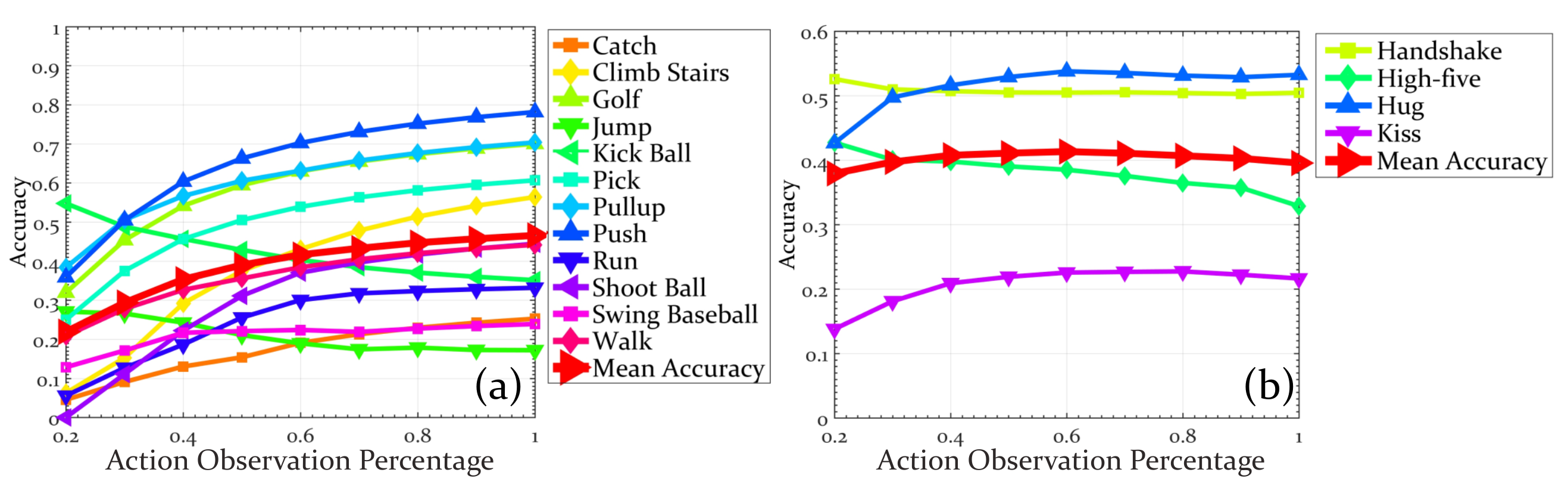}
\caption{This figure shows per-action prediction accuracy as a function of observed action (interaction) percentage for (a) sub-JHMDB and (b) TV Human Interaction datasets. The mean accuracy for all actions (interactions) is shown with bold red curve.}
\label{fig:Prediction_Per_Action}
%\vspace{-.1in}
\end{figure*}

\begin{figure*}
\centering
\includegraphics[width=1.7\columnwidth]{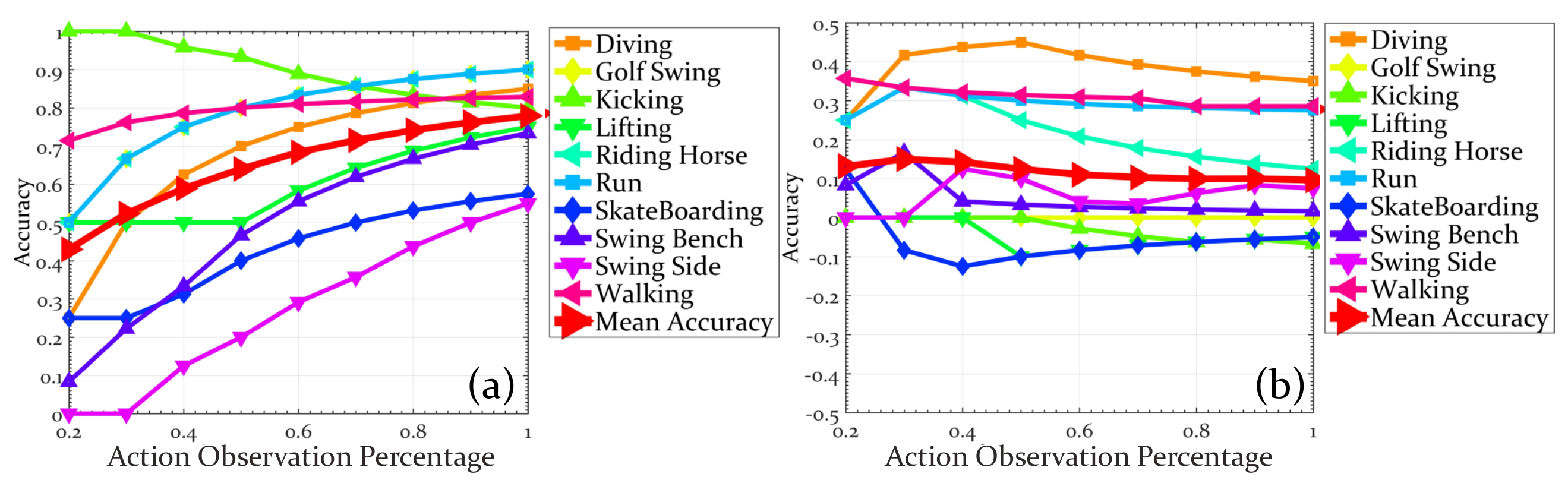}
\caption{This figure shows per-action prediction accuracy as a function of observed action percentage for UCF Sports dataset for (a) Structural SVM approach (Sec. \ref{subsec:SSVM}) and (b) and its difference with SVM and Dynamic Programming (Sec. \ref{subsec:SVM_DP}). On average, S-SVM outperforms DP-SVM.}
\label{fig:Prediction_Per_ActionUCF}
%\vspace{-.1in}
\end{figure*}

\begin{table*}
\footnotesize{
\caption{This table shows the the percentage of video observation required to achieve a prediction accuracy of 30\%. Results in the first two rows are from JHMDB, then from sub-JHMDB and the last row is from UCF Sports dataset. Actions with missing values indicate that they did not reach a prediction accuracy of 30\% until video completion.}
\begin{center}
\scalebox{.9}{
\begin{tabular}{c||cccccccccccc}
\specialrule{1.5pt}{1pt}{1pt}
\toprule
\textbf{\shortstack{JHMDB\\ Actions}} & \textit{\shortstack{Shoot \\ Ball}} & \textit{\shortstack{Shoot \\ Gun}} & \textit{Pull up} & \textit{Golf} & \textit{Clap} & \textit{\shortstack{Climb \\ Stairs}} & \textit{\shortstack{Shoot \\ Bow}} & \textit{\shortstack{Brush \\ Hair}} & \textit{Pour}  & \textit{Push} & \textit{Walk} \\

\hline
  \textit{Video (\%)} & \textit{1\%} & \textit{1\%} & \textit{16\%} & \textit{19\%} & \textit{25\%} & \textit{26\%} & \textit{28\%} & \textit{32\%} & \textit{32\%} & \textit{36\%} & \textit{36\%} \\

\midrule
%\hline
 \textbf{\shortstack{JHMDB\\ Actions}} & \textit{Sit} & \textit{\shortstack{Swing \\ Baseball}} & \textit{Run} & \textit{Stand} &  \textit{Catch}  & \textit{Jump} & \textit{Pick} & \textit{\shortstack{Kick \\ Ball}} & \textit{Throw} & \textit{Wave} \\

\hline
 \textit{Video (\%)} & \textit{40\%} & \textit{40\%} & \textit{48\%} & \textit{60\%} & - & - & - & - & - & - \\

\midrule
 \textbf{\shortstack{sub-JHMDB\\ Actions}} & \textit{\shortstack{Kick \\ Ball}} & \textit{Pullup} & \textit{Golf} & \textit{Push} & \textit{Walk} & \textit{Pick} & \textit{\shortstack{Climb \\ Stairs}} & \textit{\shortstack{Shoot \\ Ball}} &  \textit{Run} & \textit{Catch} & \textit{Jump} & \textit{\shortstack{Swing \\ Baseball}}\\

\midrule
 \textit{Video (\%)} & \textit{1\%} & \textit{17\%} & \textit{18\%} & \textit{18\%} & \textit{20\%} & \textit{24\%} & \textit{41\%} & \textit{48\%} & \textit{60\%} & - & - & -\\

\midrule
%\hline
 \textbf{\shortstack{UCF Sports \\ Actions}} & \textit{Kicking} & \textit{Lifting} & \textit{Walking} & \textit{\shortstack{Golf \\ Swing}} & \textit{\shortstack{Riding \\ Horse}} & \textit{Run} & \textit{Diving} & \textit{\shortstack{Swing \\ Bench}} & \textit{\shortstack{Skate \\ Boarding}} & \textit{\shortstack{Swing \\ Side}} \\

\hline
 \textit{Video (\%)} & \textit{1\%} & \textit{1\%} & \textit{1\%} & \textit{15\%} & \textit{15\%} & \textit{15\%} & \textit{22\%} & \textit{36\%} & \textit{37\%} & \textit{61\%} \\

\bottomrule
\specialrule{1.5pt}{1pt}{1pt}
\end{tabular}}
\end{center}
\label{table:VOPTable}
}
%\vspace{-.1in}
\end{table*}

\begin{figure*}[t]
\centering
\includegraphics[width=1.8\columnwidth]{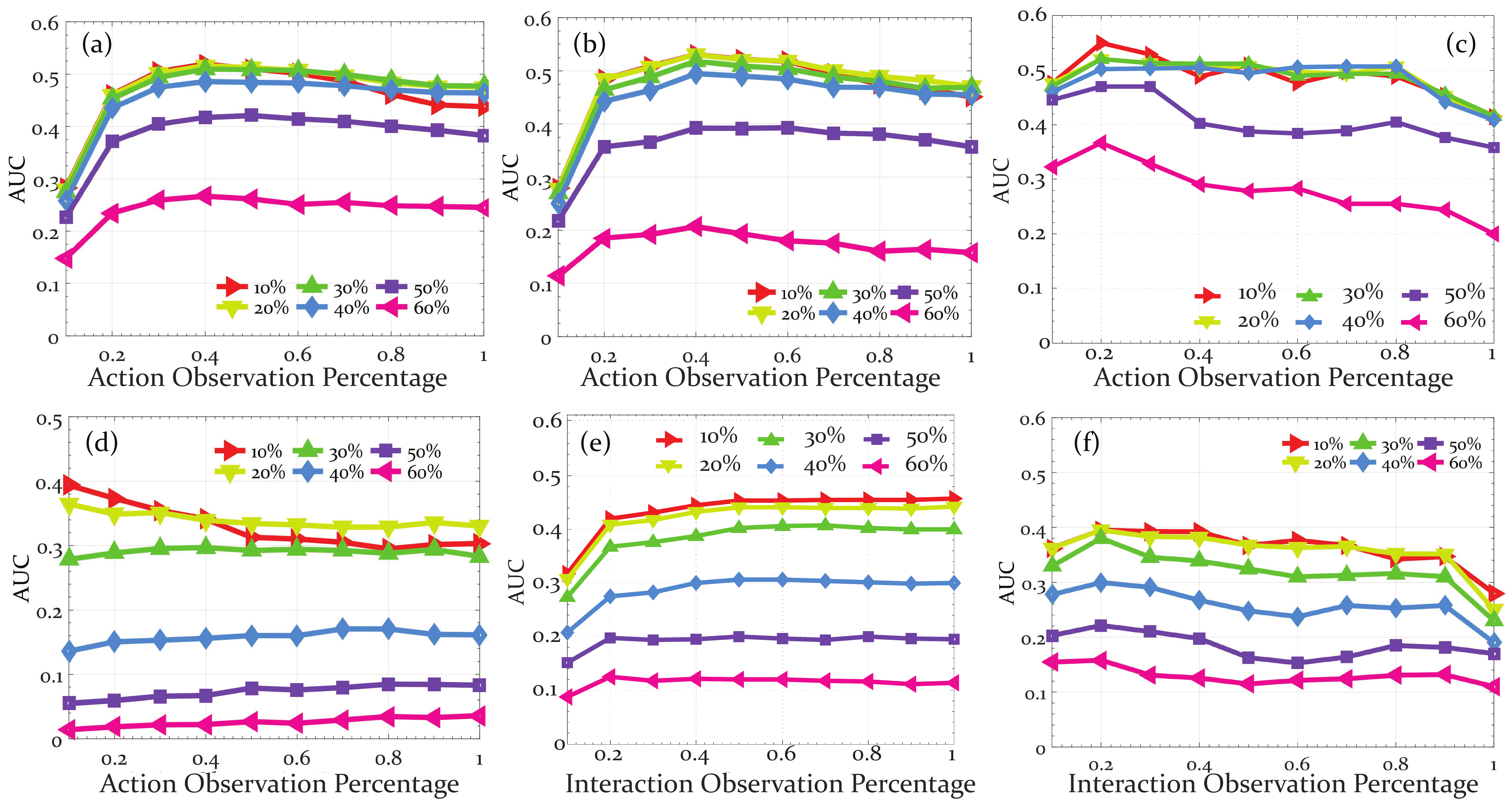}
\caption{This figure shows online action (interaction) localization performance as a function of observed action percentage on (a) JHMDB, (b) sub-JHMDB, (c) UCF-Sports, (d) MSR-II, and as a function of observed interaction percentage for (e) TV Human Interaction and (f) UT Interaction datasets. Different curves show evaluations at different overlap thresholds: $10\%$ (red), $30\%$ (green) and $60\%$ (pink).}
\label{fig:finalOnlineAUC}
\end{figure*}

\subsection{Results and Analysis}

%\smallskip
\noindent\textbf{Action (Interaction) Prediction with Time:} The prediction accuracy is evaluated with respect to the percentage of action (interaction) observed. Fig. \ref{fig:PredvsTime} shows the accuracy against time for (a) UCF Sports, (b) JHMDB, (c) sub-JHMDB, (d) MSR-II, (e) TV Human Interaction and (f) UT Interaction datasets. The results show that Structural SVM in general performs better than Binary SVM with Dynamic Programming as it learns to predict higher confidence as more action is observed. It is evident that predicting the class of an action based on partial observation is very challenging, and the accuracy of correctly predicting the action increases as more information becomes available. However, the curves for MSR-II (Fig. \ref{fig:PredvsTime}(e)) and UT Interaction (Fig. \ref{fig:PredvsTime}(g)) datasets do not reflect noticeable change as more action (interaction) is observed. This is partially due to the reason that both these datasets have very few classes (3 and 6, respectively), and there is little confusion among classes from the onset of actions. An analysis of prediction accuracy per action class is shown in Fig. \ref{fig:Prediction_Per_Action} for (a) sub-JHMDB and (b) TV Human Interaction datasets. Similarly Fig. \ref{fig:Prediction_Per_ActionUCF}(a) shows per-action results for UCF Sports. A common theme among the results of all the datasets is that actions which have actors in upright standing position are always easy to predict and localize compared to other actions. This is also visible from the curves of \textit{kicking} (UCF Sports), \textit{kick ball} (sub-JHMDB) and \textit{hand shake, high five} (TV Human Interaction) which begin with a high prediction accuracy and drop slightly as observation time period progresses, thus suggesting strong bias of classifier towards such actions (interactions). For sub-JHMDB, high prediction accuracy actions include \textit{push} and \textit{pull up}, both of which have humans in upright position making pose estimation easy, whereas \textit{jump} is the most difficult action to predict. An inspection of videos for this action reveals that most of the instances were taken from parkour exhibiting high articulation and intra-class variation. For TV Human Interaction dataset, \textit{hug} is easy to predict whereas \textit{kiss} is the most difficult due to its subtle motion and high confusion with \textit{hug}. For UCF Sports, high performing actions are \textit{kicking, walking} and \textit{running}, all upright with smooth motion of legs. For this dataset, the most difficult action is \textit{swing side} due to high articulation with most of the instances depicting swinging of the sportsperson from the very first frame with different pose at the beginning of each action instance. Finally, we also analyze the performance of DP-SVM and S-SVM in Fig. \ref{fig:Prediction_Per_ActionUCF}(b) which shows the difference of prediction accuracy DP-SVM and S-SVM. Longer duration actions such as \textit{diving, walking, running, riding horse} gain significant boost in prediction accuracy, with average performance increasing by about 12\% over the baseline DP-SVM for this dataset.

Since each action has its own predictability, we also analyze how early we can predict each action. We arbitrarily set the prediction accuracy to 30\% and show the percentage of action observation required for each action of JHMDB, sub-JHMDB and UCF Sports datasets in Table \ref{table:VOPTable}. Although we set a reasonable prediction target, certain actions do not reach such prediction accuracy even until the completion of the video. This highlights the challenging nature of online action prediction and localization.

\begin{figure*}[t]
\centering
\includegraphics[width=2\columnwidth]{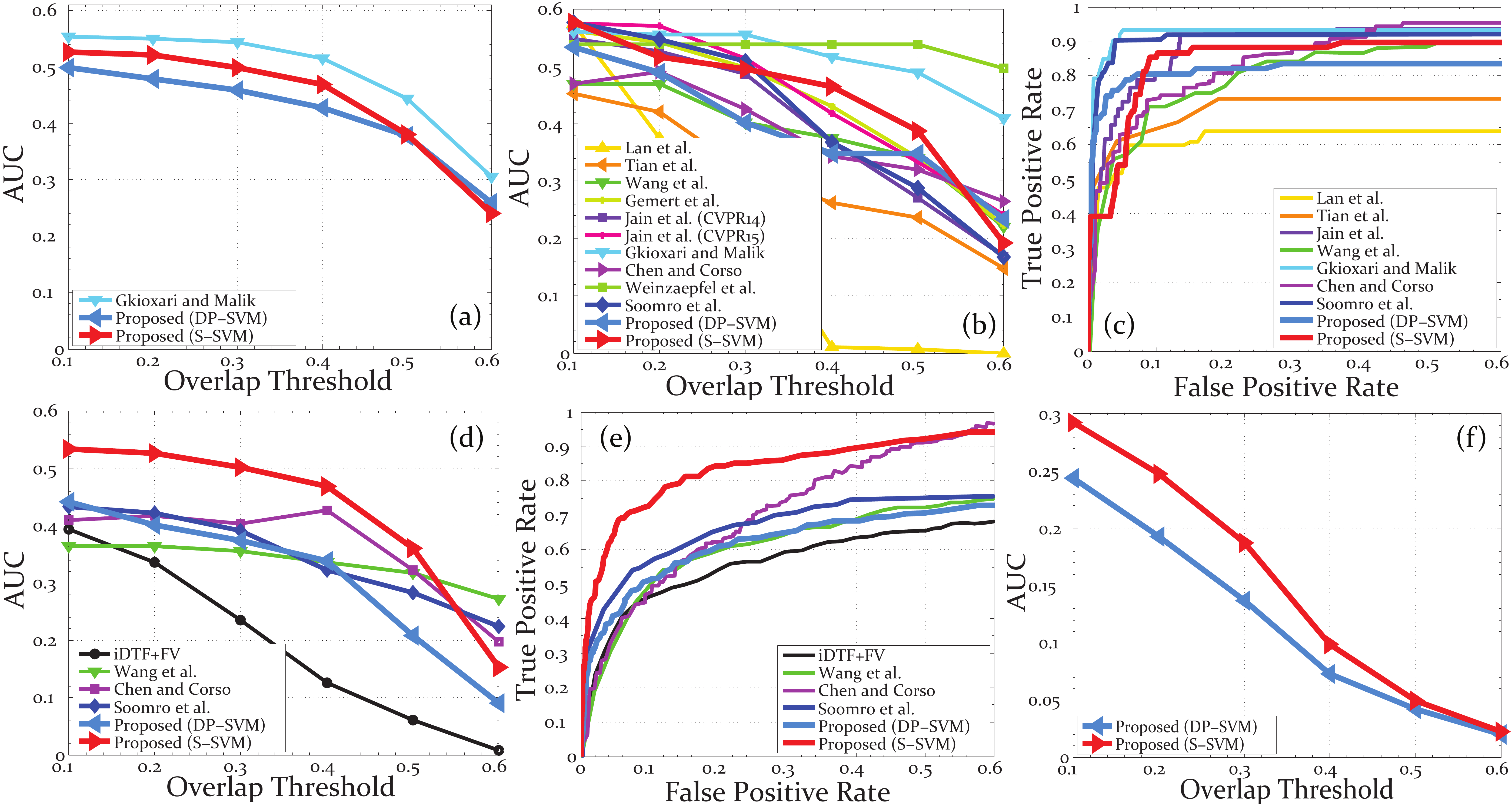}
\caption{This figure shows action localization results of the baseline Binary SVM with Dynamic Programming (DP-SVM) and Structural SVM (S-SVM) approaches, along with existing \textit{offline} methods on four action datasets (JHMDB, UCF Sports, sub-JHMDB and MSR-II). (a) shows AUC curves for JHMDB, while (b) and (c) show AUC and ROC @ $20\%$, respectively, for UCF Sports dataset. AUC and ROC @ $20\%$ overlap are shown in (d) and (e) for sub-JHMDB dataset, finally AUC for MSR-II dataset is shown in (f). The curve for S-SVM method is shown in red and DP-SVM is shown in blue, while other \textit{offline} localization methods including Lan~\etal~\cite{mori11}, Tian~\etal~\cite{sdpm13}, Wang~\etal~\cite{wang2014video}, van Gemert~\etal~\cite{van2015apt}, Jain~\etal~\cite{tubelets14}~\cite{jain201515}, Gkioxari~and~Malik~\cite{gkioxari2014finding}, Chen~and~Corso~\cite{chen2015action}, Weinzaepfel~\etal~\cite{weinzaepfel2015learning} and Soomro~\etal~\cite{soomro2015action} are shown with different colors. Despite being online, the proposed approach performs competitively overall compared to existing offline methods.}
\label{fig:AUC_ROC}
%\vspace{-.1in}
\end{figure*}

\smallskip
\noindent\textbf{Action (Interaction) Localization with Time:} To evaluate online performance, we analyze how the localization performance varies across time by computing prediction accuracy as a function of observed action (interaction) percentage. Fig. \ref{fig:finalOnlineAUC} shows the AUC against time for different overlap thresholds ($10\%-60\%$) for (a) JHMDB, (b) sub-JHMDB, (c) UCF Sports and (d) MSR-II action datasets. The AUC as a percentage of observed interaction percentage is shown for (e) TV Human Interaction and (f) UT Interaction datasets as well. We compute the AUC with time in a cumulative manner such that the accuracy at $50\%$ means localizing an action from start till one-half of the video has been observed. This gives an insight into how the overall localization performance varies as a function of time or observed percentage in testing videos. These graphs show that it is challenging to localize an action at the beginning of the video, since there is not enough discriminative motion observed by the algorithm to distinguish different actions. Furthermore, our approach first learns an appearance model from pose bounding boxes, which are improved and refined as time progresses. This improves the superpixel-based appearance confidence, which then improves the localization, and stabilizes the AUC.
The curves also show that the AUC is inversely proportional to the overlap threshold. %Therefore, AUC decreases with an increase in overlap threshold.

%\smallskip
%\noindent\textbf{Observations:}
There are two interesting observations that can be made from these graphs. First, for the JHMDB, sub-JHMDB and MSR-II datasets in Fig. \ref{fig:finalOnlineAUC}(a,b,d), the results improve initially, but then deteriorate in the middle, i.e. when the observation percentage is around $60\%$. The reason is that most of the articulation and motion happens in the middle of the video. Thus, the segments in the middle are the most difficult to localize, resulting in drop of performance. Second, the curves for UCF Sports in Fig. \ref{fig:finalOnlineAUC}(c) depict a rather unexpected behavior in the beginning, where localization improves and then suddenly worsens at around $15\%$ observation percentage. On closer inspection, we found that this is due to rapid motion in some of the actions, such as \textit{diving} and \textit{swinging (side view)}. For these actions, the initial localization is correct when the actor is stationary, but both actions have very rapid motion in the beginning, which violates the continuity constraints applicable to many other actions. This results in a drop in performance, and since this effect accumulates as observation percentage increases, the online algorithm never attains the peak again for many overlap thresholds despite observing the entire action.

\begin{figure*}[t]
\centering
\includegraphics[width=2\columnwidth]{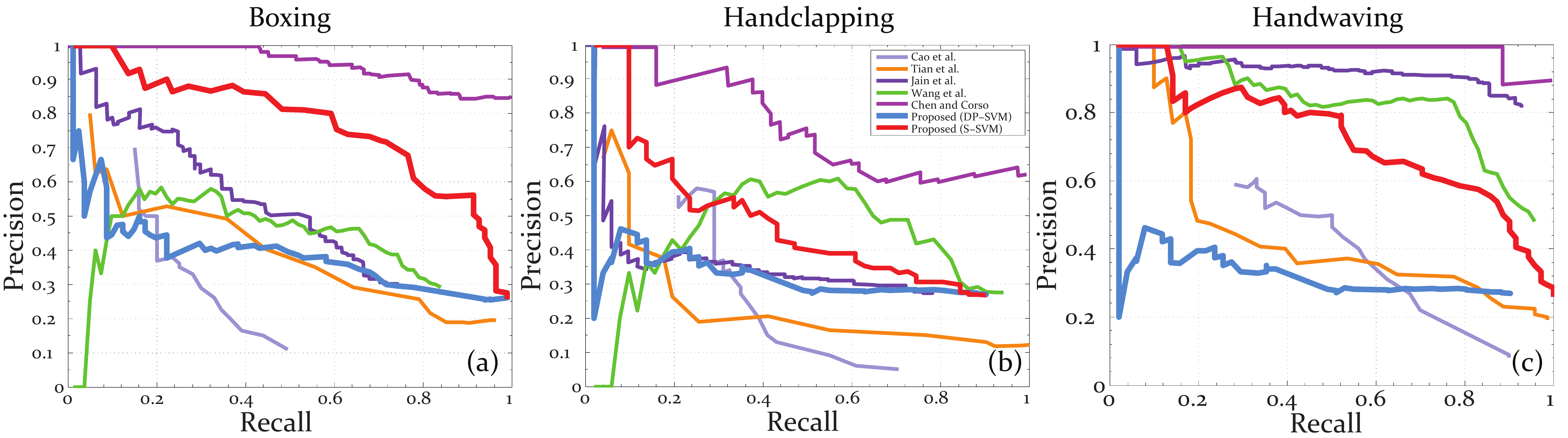}
\caption{This figure shows action localization results on MSR-II dataset. The precision/recall curves are drawn for three actions: (a) boxing, (b) Hand clapping and (c) hand waving. We perform competitive to many existing \textit{offline} methods. Red curve shows the proposed S-SVM approach, while blue curve shows the results of the baseline DP-SVM method.}
\label{fig:MSR2_PR}
%\vspace{-.1in}
\end{figure*}

\smallskip
\noindent\textbf{Comparison with Offline Methods:} We also evaluate the performance of our method against existing \textit{offline} state-of-the-art action localization methods. Fig. \ref{fig:AUC_ROC}(a) shows the results of the proposed S-SVM method, on JHMDB dataset, in red and the baseline DP-SVM in blue, while that of \cite{gkioxari2014finding} in cyan. The difference in performance is attributed to the online vs. offline nature of the methods, as well as the use of CNN features by \cite{gkioxari2014finding}. Quantitative comparison on UCF Sports using AUC and ROC @ $20\%$ is shown in Fig. \ref{fig:AUC_ROC}(b) and (c) respectively. Fig. \ref{fig:AUC_ROC} also shows the results of S-SVM and DP-SVM over all datasets where S-SVM outperforms DP-SVP highlighting the importance of structured prediction. The biggest gain in performance is visible in sub-JHMDB dataset, as shown by the AUC and ROC curves in Fig. \ref{fig:AUC_ROC}(d) and (e), where despite being online S-SVM outperforms existing state-of-the-art methods.

For MSR-II dataset, we evaluate action localization and prediction using two separate metrics with: 1) precision/recall curve to draw comparison with existing methods as shown in Fig \ref{fig:MSR2_PR} for the three different actions: (a) boxing, (b) hand clapping and (c) hand waving. 2) AUC performance is also shown in Fig. \ref{fig:AUC_ROC} (f) for consistent evaluation with other datasets. The average precision per action is presented in Table \ref{table:MSR_PR_Table}.

\begin{table}
\footnotesize{
\caption{This table shows the average precision for MSR-II dataset on three different actions: (a) Boxing, (b) Handclapping and (c) Handwaving.}
\begin{center}
\scalebox{1}{
\begin{tabular}{c||c|c|c}
\specialrule{1.5pt}{1pt}{1pt}
\toprule
\textbf{Method} & \textit{Boxing} & \textit{Handclapping} & \textit{Handwaving} \\

\hline
  \textit{Cao \etal \cite{cao2010cross}} & \textit{17.5} & \textit{13.2} & \textit{26.7} \\

\hline
  \textit{Tian \etal \cite{sdpm13}} & \textit{38.9} & \textit{23.9} & \textit{44.4} \\

\hline
  \textit{Jain \etal \cite{tubelets14}} & \textit{46.0} & \textit{31.4} & \textit{85.8} \\

\hline
  \textit{Wang \etal \cite{wang2014video}} & \textit{41.7} & \textit{50.2} & \textit{80.9} \\

\hline
  \textit{Chen and Corso \cite{chen2015action}} & \textit{94.4} & \textit{73.0} & \textit{87.7} \\

\hline
\hline
  \textit{Proposed (DP-SVM)} & \textit{37.3} & \textit{28.3} & \textit{42.9} \\

\hline
  \textit{Proposed (S-SVM)} & \textit{75.3} & \textit{43.4} & \textit{71.3} \\

\bottomrule
\specialrule{1.5pt}{1pt}{1pt}
\end{tabular}}
\end{center}
\label{table:MSR_PR_Table}
}
%\vspace{-.1in}
\end{table}

%\begin{figure}
%\centering
%\includegraphics[width=1\columnwidth]{./subJHMDB.pdf}
%\caption{This figure shows localization results of proposed method with existing methods. (a) and (b) show ROC @ $20\%$ and AUC for sub-JHMDB dataset. In these graphs, the method by \cite{wang2014video} is shown in green, iDTF+FV baseline \cite{wang2014video} in black and proposed method in red. Despite being online, the proposed method performs competitively to offline methods when evaluated for entire videos.}
%\label{fig:subJHMDB}
%\end{figure}

\begin{figure}
\centering
\includegraphics[width=1\columnwidth]{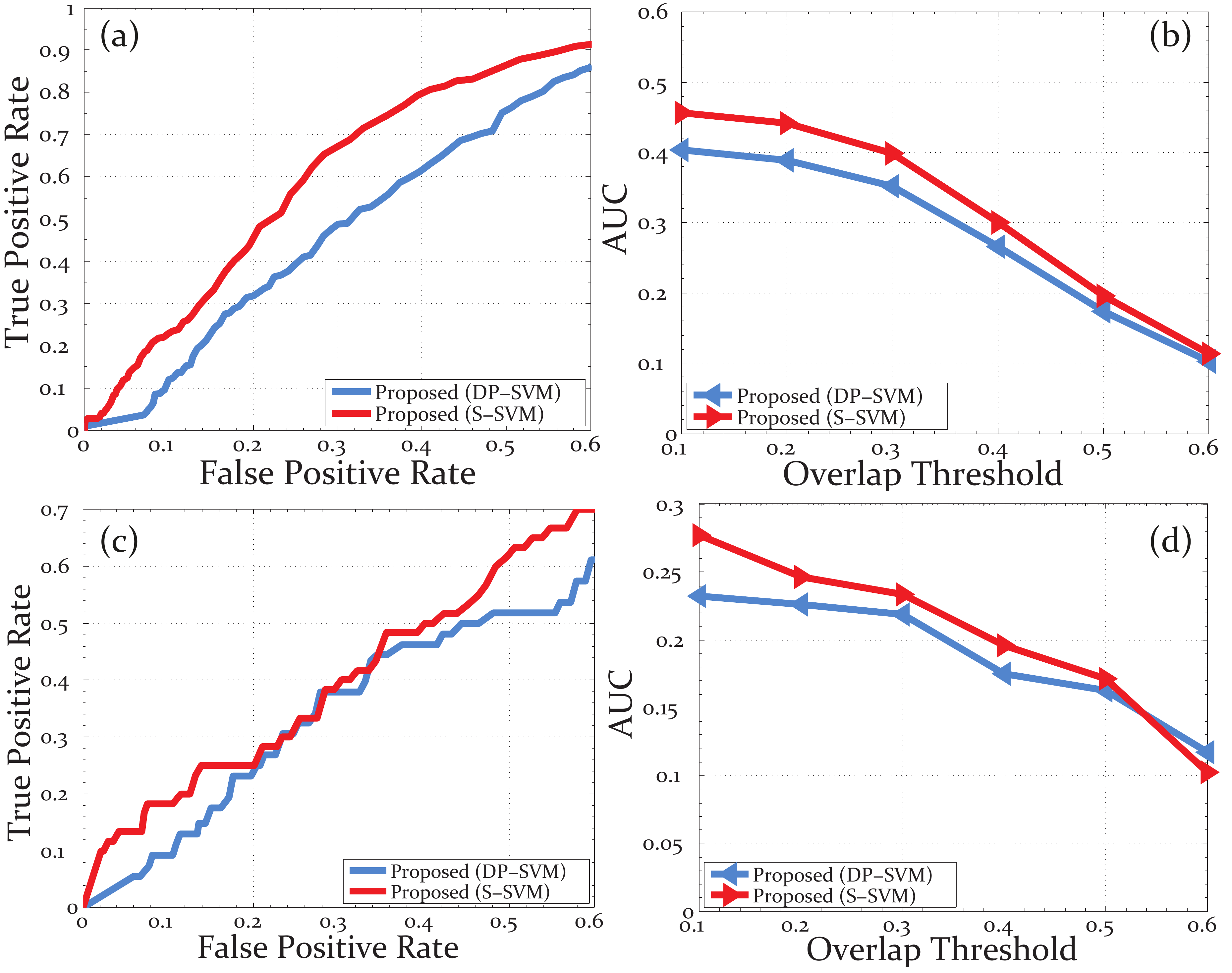}
\caption{This figure shows interaction localization results on two interaction datasets. ROC @ $20\%$ overlap and AUC curves for TV Human Interaction dataset are shown in (a) and (b), and for UT Interaction dataset in (c) and (d). In this figure, S-SVM shown in red and DP-SVM (baseline) in blue.}
\label{fig:UT_Interaction_ROC_AUC}
\end{figure}

\smallskip
Generally, interaction datasets have either been used for classification \cite{ryoo2009spatio}, activity prediction \cite{ryoo2011human} or video retrieval \cite{patron2010high}. We are the first to evaluate online localization on these datasets. To keep evaluation metrics uniform, we present our performance on localization and prediction of human interactions in Fig. \ref{fig:UT_Interaction_ROC_AUC} using ROC and AUC curves for TV Human interaction (a,b) and UT interaction (c,d).

\begin{figure}[b]
\centering
\includegraphics[width=1\columnwidth]{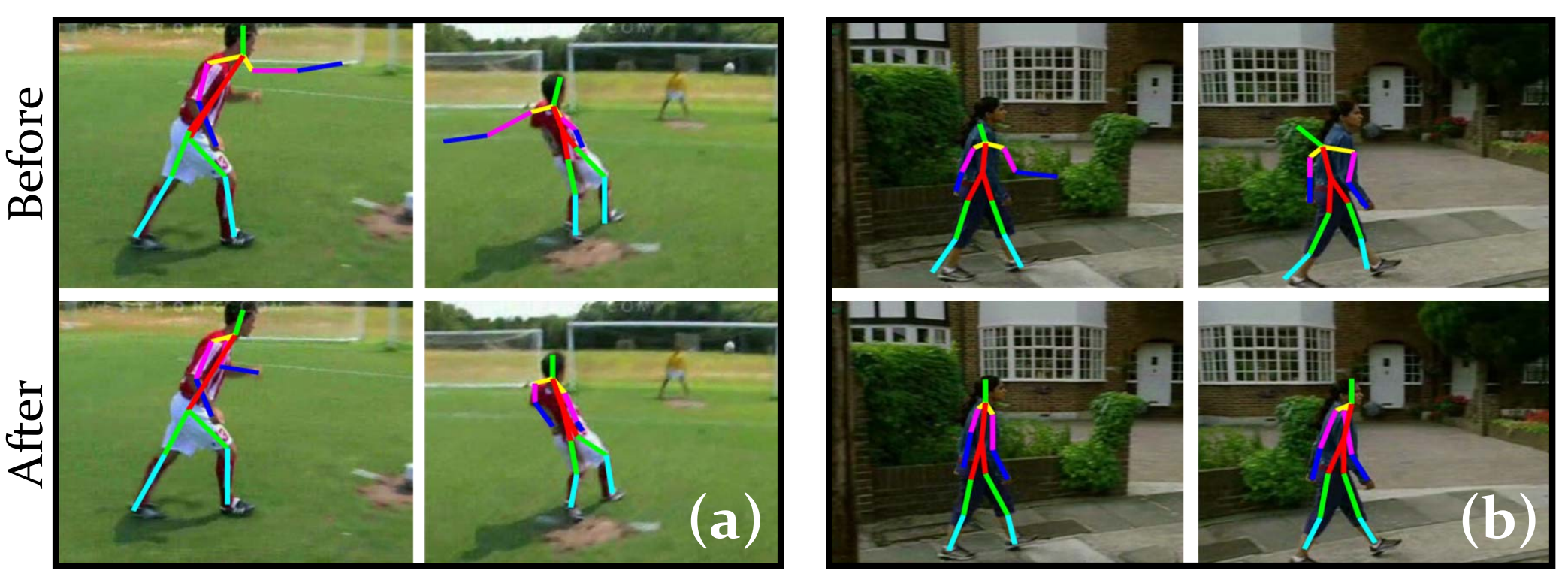}
\caption{This figure shows qualitative results for pose refinement. Results show a comparison of raw poses (top row) and refined poses (bottom row) for (a) Kicking and (b) Walking.}
\label{fig:finalQualCVPR_Pose}
%\vspace{-.1in}
\end{figure}

\begin{figure*}[t]
\centering
\includegraphics[width=2\columnwidth]{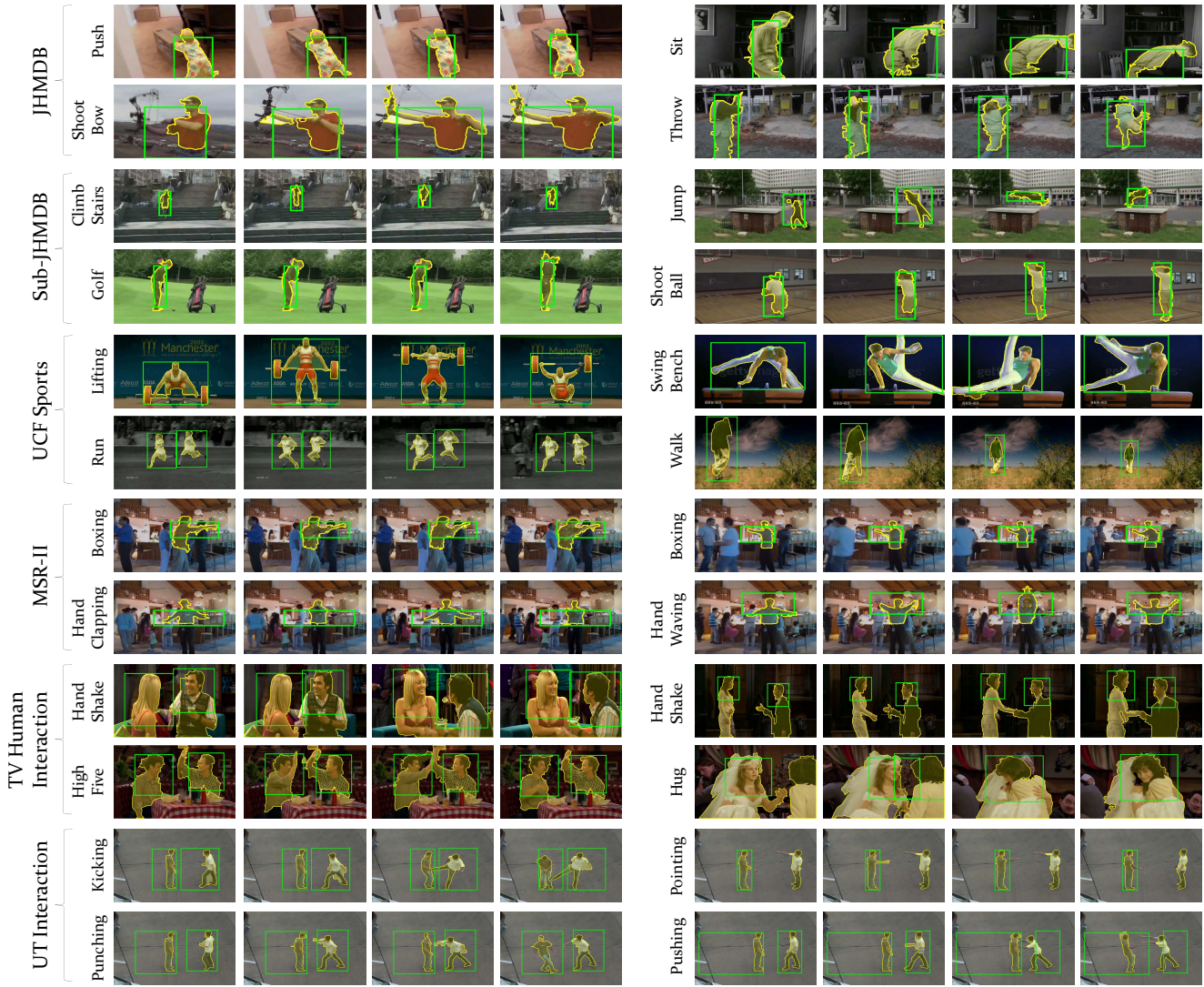}
\caption{This figure shows qualitative results of the proposed approach for all six datasets, where each action segment is shown with yellow contour and ground truth with green bounding box.}
\label{fig:finalQual}
\end{figure*}

\smallskip
\noindent\textbf{Pose Refinement:} Pose-based foreground likelihood refines poses in an iterative manner using spatio-temporal smoothness constraints. Our qualitative results in Fig. \ref{fig:finalQualCVPR_Pose} show the improvement in pose joint locations on two example videos..

\smallskip
\noindent\textbf{Action Segments:} Since we use superpixel segmentation to represent the foreground actor, our approach outputs action segments. Our qualitative results in Fig. \ref{fig:finalQual} show the fine contour of each actor (yellow) along with the ground truth (green). Using superpixels and CRF, we are able to capture the shape deformation of the actors.

\section{Conclusion}\label{sec:conclusion}

In this paper, we introduced a new prediction problem of online action and interaction localization, where the goal is to simultaneously localize and predict action (interaction) in an online manner. We presented an approach which uses representations at different granularities - from high-level poses for initialization, mid-level features for generating action tubes, and low-level features such as iDTF for action (interaction) prediction. We also refine pose estimation online using spatio-temporal constraints. The localized tubes are obtained using CRF, and prediction confidences come from the classifier. We showed that the Structural SVM (S-SVM) formulation outperforms the baseline dynamic programming with SVM (DP-SVM) hybrid. The intermediate results and analysis indicate that such an approach is capable of addressing this difficult problem, and performing competitive to some of the recent offline action localization methods.

%For future research, we plan to leverage training data to perform localization and prediction simultaneously by learning costs for superpixel merging for different A\&IA.

%\bigskip
%\noindent\textbf{Acknowledgment:} This material is based upon work supported in part by, the U.S. Army Research Laboratory, the U.S. Army Research Office under contract/grant number W911NF-14-1-0294.

%\clearpage
\bibliographystyle{IEEETran} %IEEETran
%\begin{thebibliography}{1}
\bibliography{PAMI_ActionLocPredV2}

\begin{thebibliography}{10}
\providecommand{\url}[1]{#1}
\csname url@samestyle\endcsname
\providecommand{\newblock}{\relax}
\providecommand{\bibinfo}[2]{#2}
\providecommand{\BIBentrySTDinterwordspacing}{\spaceskip=0pt\relax}
\providecommand{\BIBentryALTinterwordstretchfactor}{4}
\providecommand{\BIBentryALTinterwordspacing}{\spaceskip=\fontdimen2\font plus
\BIBentryALTinterwordstretchfactor\fontdimen3\font minus
  \fontdimen4\font\relax}
\providecommand{\BIBforeignlanguage}[2]{{%
\expandafter\ifx\csname l@#1\endcsname\relax
\typeout{** WARNING: IEEEtran.bst: No hyphenation pattern has been}%
\typeout{** loaded for the language `#1'. Using the pattern for}%
\typeout{** the default language instead.}%
\else
\language=\csname l@#1\endcsname
\fi
#2}}
\providecommand{\BIBdecl}{\relax}
\BIBdecl

\bibitem{mori11}
T.~Lan, Y.~Wang, and G.~Mori, ``Discriminative figure-centric models for joint
  action localization and recognition,'' in \emph{ICCV}, 2011.

\bibitem{wang2014video}
L.~Wang, Y.~Qiao, and X.~Tang, ``Video action detection with relational
  dynamic-poselets,'' in \emph{ECCV}, 2014.

\bibitem{yu2011fast}
G.~Yu, N.~A. Goussies, J.~Yuan, and Z.~Liu, ``Fast action detection via
  discriminative random forest voting and top-k subvolume search,'' \emph{IEEE
  Transactions on Multimedia}, vol.~13, no.~3, 2011.

\bibitem{dehghan2014improving}
A.~Dehghan, H.~Idrees, and M.~Shah, ``Improving semantic concept detection
  through the dictionary of visually-distinct elements,'' in \emph{CVPR}, 2014.

\bibitem{soomro2016predicting}
K.~Soomro, H.~Idrees, and M.~Shah, ``Predicting the where and what of actors
  and actions through online action localization,'' in \emph{CVPR}, 2016.

\bibitem{sdpm13}
Y.~Tian, R.~Sukthankar, and M.~Shah, ``Spatiotemporal deformable part models
  for action detection,'' in \emph{CVPR}, 2013.

\bibitem{tubelets14}
M.~Jain, J.~Gemert, H.~Jegou, P.~Bouthemy, and C.~Snoek, ``Action localization
  with tubelets from motion,'' in \emph{CVPR}, 2014.

\bibitem{soomro2015action}
K.~Soomro, H.~Idrees, and M.~Shah, ``Action localization in videos through
  context walk,'' in \emph{ICCV}, 2015.

\bibitem{oneata2014efficient}
D.~Oneata, J.~Verbeek, and C.~Schmid, ``Efficient action localization with
  approximately normalized fisher vectors,'' in \emph{CVPR}, 2014.

\bibitem{oneata2014spatio}
D.~Oneata, J.~Revaud, J.~Verbeek, and C.~Schmid, ``Spatio-temporal object
  detection proposals,'' in \emph{ECCV}, 2014.

\bibitem{hoai2014talking}
M.~Hoai and A.~Zisserman, ``Talking heads: Detecting humans and recognizing
  their interactions,'' in \emph{Proceedings of the IEEE Conference on Computer
  Vision and Pattern Recognition}, 2014, pp. 875--882.

\bibitem{kong2012learning}
Y.~Kong, Y.~Jia, and Y.~Fu, ``Learning human interaction by interactive
  phrases,'' in \emph{European Conference on Computer Vision}.\hskip 1em plus
  0.5em minus 0.4em\relax Springer, 2012, pp. 300--313.

\bibitem{patron2010high}
A.~Patron-Perez, M.~Marszalek, A.~Zisserman, and I.~D. Reid, ``High five:
  Recognising human interactions in tv shows.'' in \emph{BMVC}, vol.~1.\hskip
  1em plus 0.5em minus 0.4em\relax Citeseer, 2010, p.~2.

\bibitem{patron2012structured}
A.~Patron-Perez, M.~Marszalek, I.~Reid, and A.~Zisserman, ``Structured learning
  of human interactions in tv shows,'' \emph{IEEE Transactions on Pattern
  Analysis and Machine Intelligence}, vol.~34, no.~12, pp. 2441--2453, 2012.

\bibitem{ryoo2009spatio}
M.~S. Ryoo and J.~K. Aggarwal, ``Spatio-temporal relationship match: Video
  structure comparison for recognition of complex human activities,'' in
  \emph{2009 IEEE 12th international conference on computer vision}.\hskip 1em
  plus 0.5em minus 0.4em\relax IEEE, 2009, pp. 1593--1600.

\bibitem{li2014prediction}
K.~Li and Y.~Fu, ``Prediction of human activity by discovering temporal
  sequence patterns,'' \emph{IEEE TPAMI}, vol.~36, no.~8, 2014.

\bibitem{lan2014hierarchical}
T.~Lan, T.-C. Chen, and S.~Savarese, ``A hierarchical representation for future
  action prediction,'' in \emph{ECCV}, 2014.

\bibitem{ryoo2011human}
M.~Ryoo, ``Human activity prediction: Early recognition of ongoing activities
  from streaming videos,'' in \emph{ICCV}, 2011.

\bibitem{kong2014discriminative}
Y.~Kong, D.~Kit, and Y.~Fu, ``A discriminative model with multiple temporal
  scales for action prediction,'' in \emph{ECCV}, 2014.

\bibitem{zhao2013online}
X.~Zhao, X.~Li, C.~Pang, X.~Zhu, and Q.~Z. Sheng, ``Online human gesture
  recognition from motion data streams,'' in \emph{ACM MM}, 2013.

\bibitem{hoai2014max}
M.~Hoai and F.~De~la Torre, ``Max-margin early event detectors,'' \emph{IJCV},
  vol. 107, no.~2, 2014.

\bibitem{yu2012predicting}
G.~Yu, J.~Yuan, and Z.~Liu, ``Predicting human activities using spatio-temporal
  structure of interest points,'' in \emph{ACM MM}, 2012.

\bibitem{cao2013recognize}
Y.~Cao, D.~Barrett, A.~Barbu, S.~Narayanaswamy, H.~Yu, A.~Michaux, Y.~Lin,
  S.~Dickinson, J.~M. Siskind, and S.~Wang, ``Recognize human activities from
  partially observed videos,'' in \emph{CVPR}, 2013.

\bibitem{huang2014action}
D.-A. Huang and K.~M. Kitani, ``Action-reaction: Forecasting the dynamics of
  human interaction,'' in \emph{European Conference on Computer Vision}.\hskip
  1em plus 0.5em minus 0.4em\relax Springer, 2014.

\bibitem{de2016online}
R.~De~Geest, E.~Gavves, A.~Ghodrati, Z.~Li, C.~Snoek, and T.~Tuytelaars,
  ``Online action detection,'' in \emph{ECCV}, 2016.

\bibitem{li2016online}
Y.~Li, C.~Lan, J.~Xing, W.~Zeng, C.~Yuan, and J.~Liu, ``Online human action
  detection using joint classification-regression recurrent neural networks,''
  in \emph{ECCV}, 2016.

\bibitem{xie2011unified}
Y.~Xie, H.~Chang, Z.~Li, L.~Liang, X.~Chen, and D.~Zhao, ``A unified framework
  for locating and recognizing human actions,'' in \emph{CVPR}, 2011.

\bibitem{hu2009action}
Y.~Hu, L.~Cao, F.~Lv, S.~Yan, Y.~Gong, and T.~S. Huang, ``Action detection in
  complex scenes with spatial and temporal ambiguities,'' in \emph{ICCV}, 2009.

\bibitem{desai2012detecting}
C.~Desai and D.~Ramanan, ``Detecting actions, poses, and objects with
  relational phraselets,'' in \emph{ECCV}, 2012.

\bibitem{jain201515}
M.~Jain, J.~C. van Gemert, and C.~G. Snoek, ``What do 15,000 object categories
  tell us about classifying and localizing actions?'' in \emph{CVPR}, 2015.

\bibitem{rota2015real}
P.~Rota, N.~Conci, N.~Sebe, and J.~M. Rehg, ``Real-life violent social
  interaction detection,'' in \emph{Image Processing (ICIP), 2015 IEEE
  International Conference on}.\hskip 1em plus 0.5em minus 0.4em\relax IEEE,
  2015, pp. 3456--3460.

\bibitem{kong2014modeling}
Y.~Kong and Y.~Fu, ``Modeling supporting regions for close human interaction
  recognition,'' in \emph{European Conference on Computer Vision}.\hskip 1em
  plus 0.5em minus 0.4em\relax Springer, 2014, pp. 29--44.

\bibitem{yuan2011discriminative}
J.~Yuan, Z.~Liu, and Y.~Wu, ``Discriminative video pattern search for efficient
  action detection,'' \emph{IEEE TPAMI}, vol.~33, no.~9, 2011.

\bibitem{zhou2015learning}
Z.~Zhou, F.~Shi, and W.~Wu, ``Learning spatial and temporal extents of human
  actions for action detection,'' \emph{IEEE Transactions on Multimedia},
  vol.~17, no.~4, 2015.

\bibitem{felzenszwalb2010object}
P.~F. Felzenszwalb, R.~B. Girshick, D.~McAllester, and D.~Ramanan, ``Object
  detection with discriminatively trained part-based models,'' \emph{IEEE
  TPAMI}, vol.~32, no.~9, 2010.

\bibitem{tran2012max}
D.~Tran and J.~Yuan, ``Max-margin structured output regression for
  spatio-temporal action localization,'' in \emph{NIPS}, 2012.

\bibitem{ma2013action}
S.~Ma, J.~Zhang, N.~Ikizler-Cinbis, and S.~Sclaroff, ``Action recognition and
  localization by hierarchical space-time segments,'' in \emph{ICCV}, 2013.

\bibitem{yu2015fast}
G.~Yu and J.~Yuan, ``Fast action proposals for human action detection and
  search,'' in \emph{CVPR}, 2015.

\bibitem{uijlings2013selective}
J.~R. Uijlings, K.~E. van~de Sande, T.~Gevers, and A.~W. Smeulders, ``Selective
  search for object recognition,'' \emph{IJCV}, vol. 104, no.~2, 2013.

\bibitem{gkioxari2014finding}
G.~Gkioxari and J.~Malik, ``Finding action tubes,'' in \emph{CVPR}, 2015.

\bibitem{chen2014actionness}
W.~Chen, C.~Xiong, R.~Xu, and J.~J. Corso, ``Actionness ranking with lattice
  conditional ordinal random fields,'' in \emph{CVPR}, 2014.

\bibitem{kong2014interactive}
Y.~Kong, Y.~Jia, and Y.~Fu, ``Interactive phrases: Semantic descriptions for
  human interaction recognition,'' \emph{IEEE transactions on pattern analysis
  and machine intelligence}, vol.~36, no.~9, pp. 1775--1788, 2014.

\bibitem{wu2013human}
J.~Wu, F.~Chen, and D.~Hu, ``Human interaction recognition by spatial structure
  models,'' in \emph{International Conference on Intelligent Science and Big
  Data Engineering}.\hskip 1em plus 0.5em minus 0.4em\relax Springer, 2013, pp.
  216--222.

\bibitem{yeung2016end}
S.~Yeung, O.~Russakovsky, G.~Mori, and L.~Fei-Fei, ``End-to-end learning of
  action detection from frame glimpses in videos,'' in \emph{CVPR}, 2016.

\bibitem{shou2016temporal}
Z.~Shou, D.~Wang, and S.-F. Chang, ``Temporal action localization in untrimmed
  videos via multi-stage cnns,'' in \emph{CVPR}, 2016.

\bibitem{heilbron2016fast}
F.~C. Heilbron, J.~C. Niebles, and B.~Ghanem, ``Fast temporal activity
  proposals for efficient detection of human actions in untrimmed videos,'' in
  \emph{CVPR}, 2016.

\bibitem{richard2016temporal}
A.~Richard and J.~Gall, ``Temporal action detection using a statistical
  language model,'' in \emph{CVPR}, 2016.

\bibitem{wang2013action}
H.~Wang and C.~Schmid, ``Action recognition with improved trajectories,'' in
  \emph{ICCV}, 2013.

\bibitem{wang2015action}
L.~Wang, Y.~Qiao, and X.~Tang, ``Action recognition with trajectory-pooled
  deep-convolutional descriptors,'' \emph{arXiv preprint arXiv:1505.04868},
  2015.

\bibitem{maji2011action}
S.~Majiwa, L.~Bourdev, and J.~Malik, ``Action recognition from a distributed
  representation of pose and appearance,'' in \emph{CVPR}, 2011.

\bibitem{xu2012combining}
R.~Xu, P.~Agarwal, S.~Kumar, V.~N. Krovi, and J.~J. Corso, ``Combining skeletal
  pose with local motion for human activity recognition,'' in \emph{Articulated
  Motion and Deformable Objects}, 2012.

\bibitem{yang2011articulated}
Y.~Yang and D.~Ramanan, ``Articulated pose estimation with flexible
  mixtures-of-parts,'' in \emph{CVPR}, 2011.

\bibitem{wang2013approach}
C.~Wang, Y.~Wang, and A.~L. Yuille, ``An approach to pose-based action
  recognition,'' in \emph{CVPR}, 2013.

\bibitem{raptis2013poselet}
M.~Raptis and L.~Sigal, ``Poselet key-framing: A model for human activity
  recognition,'' in \emph{CVPR}, 2013.

\bibitem{xiaohan2015joint}
B.~Xiaohan~Nie, C.~Xiong, and S.-C. Zhu, ``Joint action recognition and pose
  estimation from video,'' in \emph{CVPR}, 2015.

\bibitem{pirsiavash2014assessing}
H.~Pirsiavash, C.~Vondrick, and A.~Torralba, ``Assessing the quality of
  actions,'' in \emph{ECCV}, 2014.

\bibitem{vahdat2011discriminative}
A.~Vahdat, B.~Gao, M.~Ranjbar, and G.~Mori, ``A discriminative key pose
  sequence model for recognizing human interactions,'' in \emph{Computer Vision
  Workshops (ICCV Workshops), 2011 IEEE International Conference on}.\hskip 1em
  plus 0.5em minus 0.4em\relax IEEE, 2011, pp. 1729--1736.

\bibitem{wei2016convolutional}
S.-E. Wei, V.~Ramakrishna, T.~Kanade, and Y.~Sheikh, ``Convolutional pose
  machines,'' in \emph{CVPR}, 2016.

\bibitem{achanta2012slic}
R.~Achanta, A.~Shaji, K.~Smith, A.~Lucchi, P.~Fua, and S.~Susstrunk, ``Slic
  superpixels compared to state-of-the-art superpixel methods,'' \emph{IEEE
  TPAMI}, vol.~34, no.~11, 2012.

\bibitem{jhuang2013towards}
H.~Jhuang, J.~Gall, S.~Zuffi, C.~Schmid, and M.~J. Black, ``Towards
  understanding action recognition,'' in \emph{ICCV}, 2013.

\bibitem{kuehne2011hmdb}
H.~Kuehne, H.~Jhuang, E.~Garrote, T.~Poggio, and T.~Serre, ``Hmdb: a large
  video database for human motion recognition,'' in \emph{ICCV}, 2011.

\bibitem{mikel2008}
M.~Rodriguez, A.~Javed, and M.~Shah, ``Action mach: a spatio-temporal maximum
  average correlation height filter for action recognition,'' in \emph{CVPR},
  2008.

\bibitem{soomro2014action}
K.~Soomro and A.~R. Zamir, ``Action recognition in realistic sports videos,''
  in \emph{Computer Vision in Sports}.\hskip 1em plus 0.5em minus 0.4em\relax
  Springer, 2014, pp. 181--208.

\bibitem{schuldt2004recognizing}
C.~Sch{\"u}ldt, I.~Laptev, and B.~Caputo, ``Recognizing human actions: a local
  svm approach,'' in \emph{ICPR}, 2004.

\bibitem{UT-Interaction-Data}
M.~S. Ryoo and J.~K. Aggarwal, ``{UT}-{I}nteraction {D}ataset, {ICPR} contest
  on {S}emantic {D}escription of {H}uman {A}ctivities ({SDHA}),''
  http://cvrc.ece.utexas.edu/SDHA2010/Human\_Interaction.html, 2010.

\bibitem{van2015apt}
J.~C. van Gemert, M.~Jain, E.~Gati, and C.~G. Snoek, ``Apt: Action localization
  psroposals from dense trajectories,'' in \emph{BMVC}, vol.~2, 2015, p.~4.

\bibitem{chen2015action}
W.~Chen and J.~J. Corso, ``Action detection by implicit intentional motion
  clustering,'' in \emph{Proceedings of the IEEE International Conference on
  Computer Vision}, 2015.

\bibitem{weinzaepfel2015learning}
P.~Weinzaepfel, Z.~Harchaoui, and C.~Schmid, ``Learning to track for
  spatio-temporal action localization,'' in \emph{Proceedings of the IEEE
  International Conference on Computer Vision}, 2015.

\bibitem{cao2010cross}
L.~Cao, Z.~Liu, and T.~S. Huang, ``Cross-dataset action detection,'' in
  \emph{CVPR}, 2010.

\end{thebibliography}

%\end{thebibliography}

\begin{IEEEbiography}[{\includegraphics[width=1in,height=1.25in,clip,keepaspectratio]{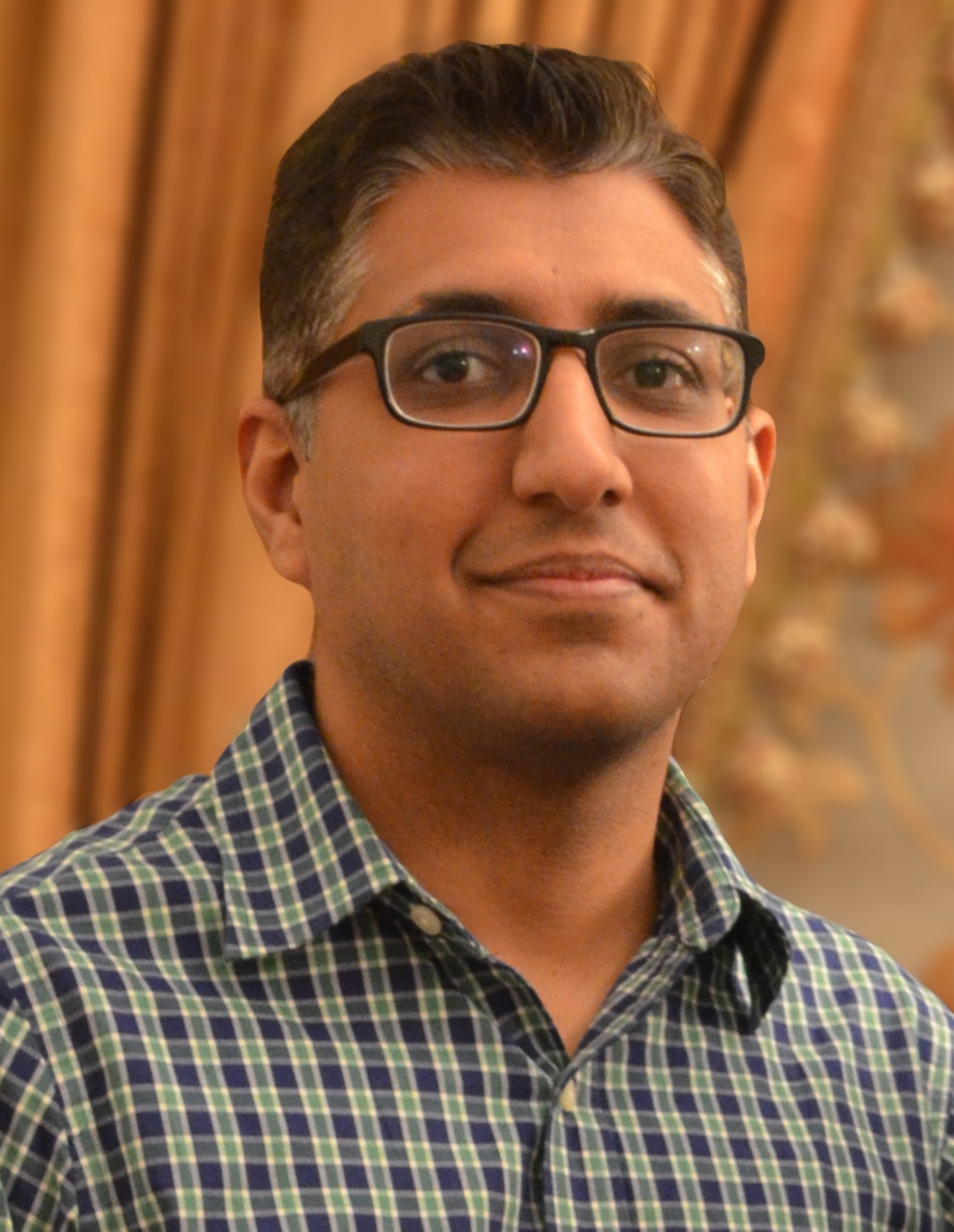}}]{Khurram Soomro}
received his B.Sc and M.Sc degrees in Computer Engineering from Lahore University of Management Sciences, Lahore, Pakistan, in 2007 and 2011, respectively. He joined Center for Research in Computer Vision at University of Central Florida in 2011, where he is currently pursuing his Ph.D degree in Computer Vision. His research interests include Action Recognition and Localization, Human Detection, Visual Surveillance and Tracking, and Sports Analytics. He is a member of the IEEE and Upsilon Pi Epsilon (UPE) Honor Society.
\end{IEEEbiography}

\begin{IEEEbiography}[{\includegraphics[width=1in,height=1.25in,clip,keepaspectratio]{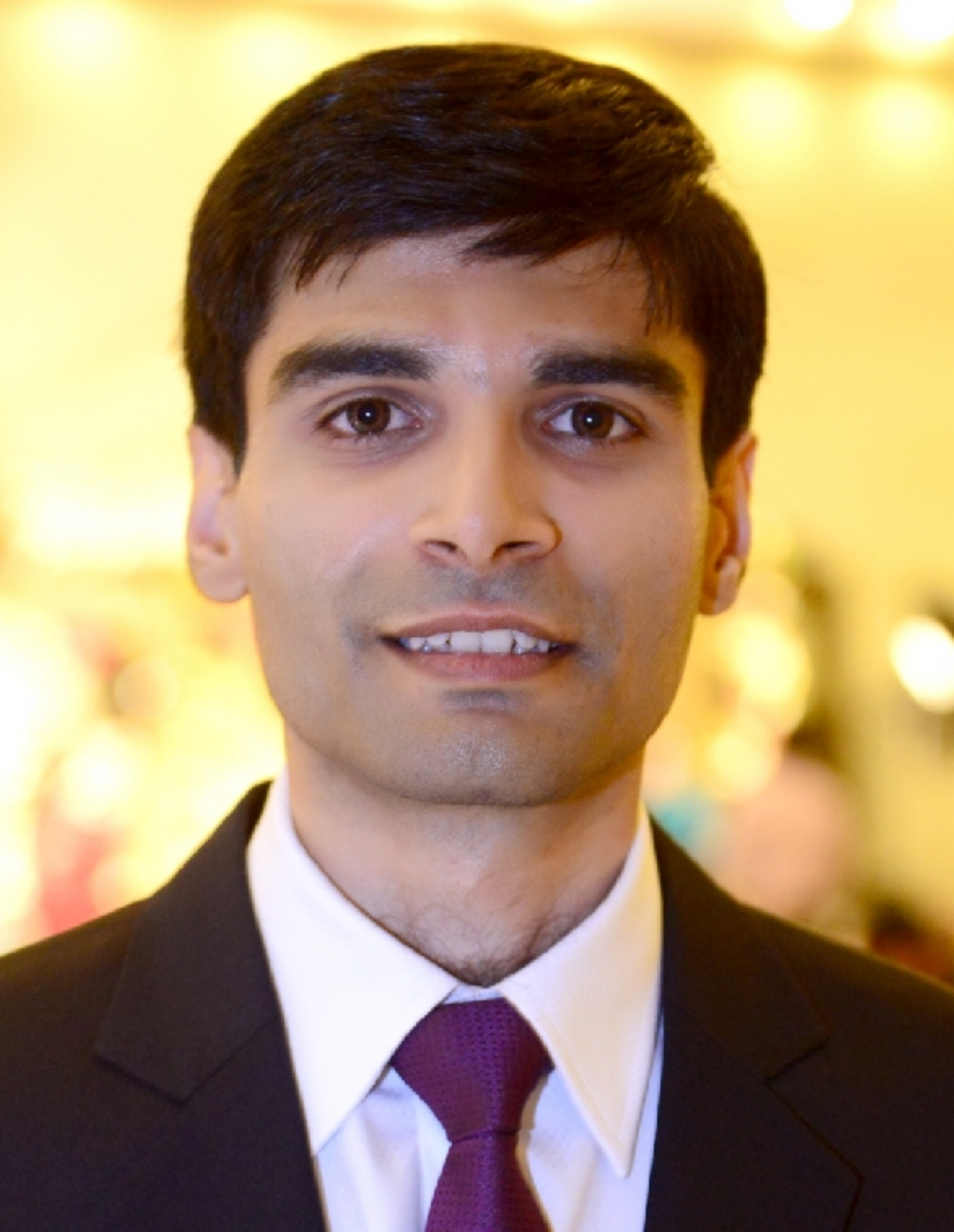}}]{Haroon Idrees}
is a Postdoctoral Associate at the Center for Research in Computer Vision at University of Central Florida. He has published several papers in conferences and journals such as CVPR, ICCV, ECCV, Journal of Image and Vision Computing, Computer Vision and Image Understanding, and IEEE Transactions on Pattern Analysis and Machine Intelligence. His research interests include crowd analysis, action recognition and localization, object detection, visual tracking, multi-camera and airborne surveillance, and multimedia content analysis. He received the BSc (Hons) degree in Computer Engineering from the Lahore University of Management Sciences, Pakistan in 2007, and the PhD degree in Computer Science from the University of Central Florida in 2014.
\end{IEEEbiography}

\begin{IEEEbiography}[{\includegraphics[width=1in,height=1.25in,clip,keepaspectratio]{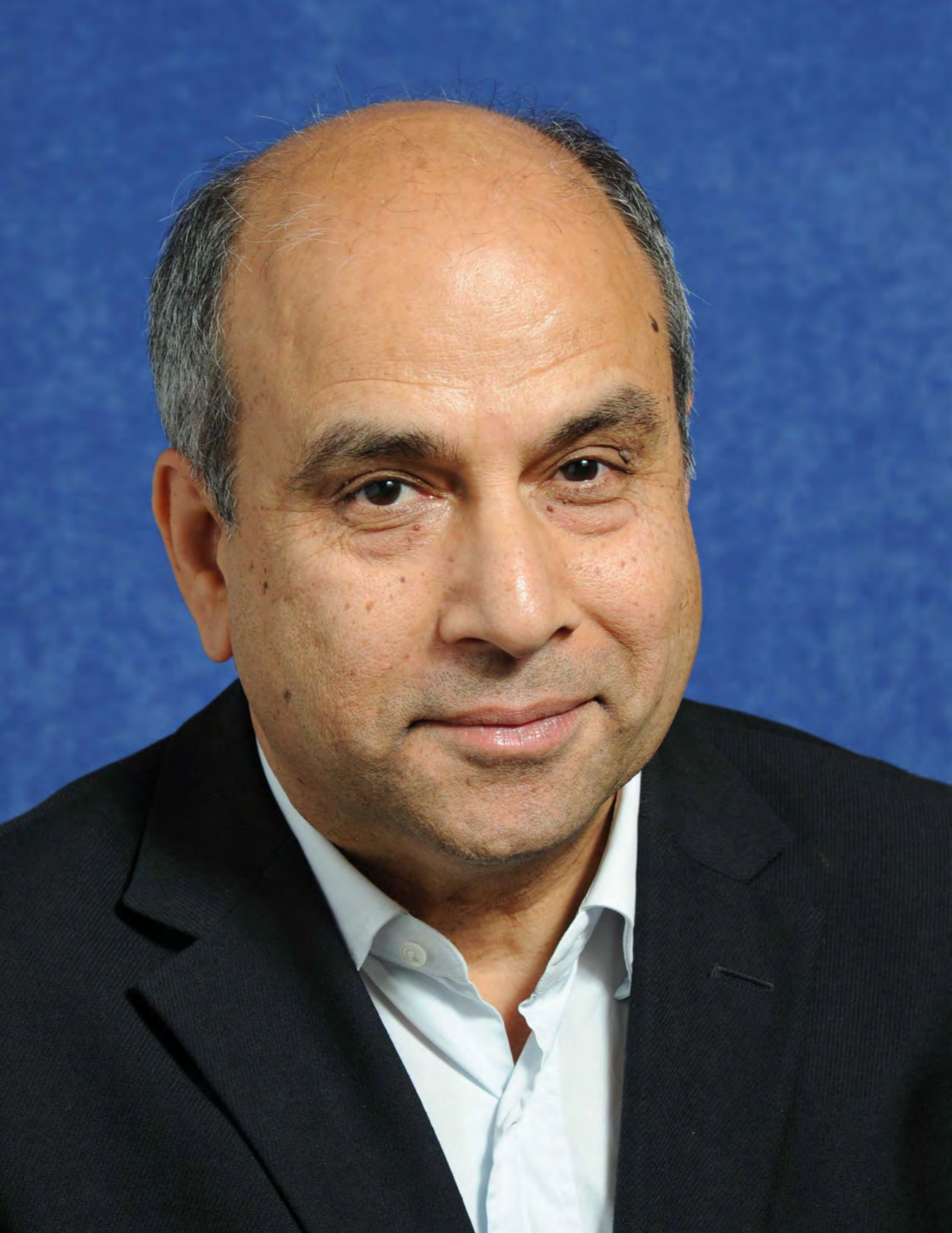}}]{Mubarak Shah,}
the Trustee chair professor of computer science, is the founding director of the Center for Research in Computer Vision at the University of Central Florida (UCF). He is an editor of an international book series on video computing, editor-in-chief of Machine Vision and Applications journal, and an associate editor of ACM Computing Surveys journal. He was the program cochair of the IEEE Conference on Computer Vision and Pattern Recognition (CVPR) in 2008, an associate editor of the IEEE Transactions on Pattern Analysis and Machine Intelligence, and a guest editor of the special issue of the International Journal of Computer Vision on Video Computing. His research interests include video surveillance, visual tracking, human activity recognition, visual analysis of crowded scenes, video registration, UAV video analysis, and so on. He is an ACM distinguished speaker. He was an IEEE distinguished visitor speaker for 1997-2000 and received the IEEE Outstanding Engineering Educator Award in 1997. In 2006, he was awarded a Pegasus Professor Award, the highest award at UCF. He received the Harris Corporation's Engineering Achievement Award in 1999, TOKTEN awards from UNDP in 1995, 1997, and 2000, Teaching Incentive Program Award in 1995 and 2003, Research Incentive Award in 2003 and 2009, Millionaire's Club Awards in 2005 and 2006, University Distinguished Researcher Award in 2007, Honorable mention for the ICCV 2005 Where Am I? Challenge Problem, and was nominated for the Best Paper Award at the ACM Multimedia Conference in 2005. He is a fellow of the IEEE, AAAS, IAPR, and SPIE.
\end{IEEEbiography}

\end{document}